%% file: main.tex
\documentclass{ieeetj}

\usepackage{textcomp}
\usepackage{tabularray}
\usepackage{makecell}

\include{latexGoodPractices/preamble}

\definecolor{asphalt}{HTML}{808080}
\definecolor{ice}{HTML}{0000FF}
\definecolor{gravel}{HTML}{FFA500}
\definecolor{grass}{HTML}{008000}
\definecolor{sand}{HTML}{FF4500}
\definecolor{mud}{HTML}{B8860B}
\definecolor{tile}{HTML}{F08080}

\usepackage{algorithm,algorithmic}
\def\BibTeX{{\rm B\kern-.05em{\sc i\kern-.025em b}\kern-.08em
    T\kern-.1667em\lower.7ex\hbox{E}\kern-.125emX}}
\AtBeginDocument{\definecolor{tmlcncolor}{cmyk}{0.93,0.59,0.15,0.02}\definecolor{NavyBlue}{RGB}{0,86,125}}

\def\OJlogo{\vspace{-4pt}$<$Society logo(s) and publication title will appear here.$>$}
\def\seclogo{\vspace{10pt}$<$Society logo(s) and publication title will appear here.$>$}

\def\authorrefmark#1{\ensuremath{^{\textbf{#1}}}}

\begin{document}

\newcommand{\algorithmautorefname}{Algorithm} 


\acrodef{UGV}{uncrewed ground vehicle} 
\acrodef{GP}{Gaussian process}
\acrodef{SSMR}{skid-steering mobile robot}
\acrodef{ATV}{all-terrain vehicle}
\acrodef{MPPI}{model predictive path integral}

\acrodef{SLAM}{simultaneous localization and mapping}
\acrodef{SOTA}{state-of-the-art}
\acrodef{SSMR}{skid-steering mobile robot}
\acrodef{AMR}{Ackermann mobile robot}
\acrodef{UGV}{uncrewed ground vehicle} 
\acrodef{DOF}{degrees of freedom}
\acrodef{IDD}{ideal differential-drive}
\acrodef{ICR}{instantaneous center or rotation}
\acrodef{RTK}{Realtime Kinematics}
\acrodef{GNSS}{Global Navigation Satellite System}
\acrodef{ROC}{radius of curvature}
\acrodef{IMU}{inertial measurement unit}
\acrodef{MPC}{model predictive control}
\acrodef{GP}{Gaussian process}
\acrodef{BLR}{Bayesian linear regression}
\acrodef{IPEM}{integrated prediction error minimization}
\acrodef{MLP}{multilayer perceptron}
\acrodef{ICP}{iterative closest point}
\acrodef{MRMSE}{multi-step root mean squared error}
\acrodef{T-MRMSE}{translational multi-step root mean squared error}
\acrodef{R-MRMSE}{rotational multi-step root mean squared error}
\acrodef{M-Z-score}{multi-step Z-score}
\acrodef{DRIVE}{Data-driven Robot Input Vector Exploration}
\acrodef{VISTA}{Vehicle Input Space Training Assistant}

\newcommand{\DOUGHNUTCALIB}{\ac{DRIVE}\xspace}
\newcommand{\ICRBASED}{\ac{ICR}-based\xspace} 
\newcommand{\DIMPARAMS}{\bm k}
\newcommand{\HORDUR}{h_d}
\newcommand{\WINSIZE}{h}
\newcommand{\TRAINDATA}{$\mathcal{D}$\xspace}
\newcommand{\DOUGHNUTDATA}{$\mathcal{D}_D$}
\newcommand{\HUMANDATA}{$\mathcal{D}_H$}
\newcommand{\PREDSTATE}{\,^\mapf\!\hat{\bm q}}
\newcommand{\MEASSTATE}{\,^\mapf\bm q}
\newcommand{\MARMOTTE}{HD2\xspace}
\newcommand{\CMDBODYVEL}{^{\robotf}\bm{f}}
\newcommand{\SLIPBODYVEL}{^{\robotf}\bm{g}}
\newcommand{\OBSERVEDSLIP}{\bm{g}}
\newcommand{\INPUTVECTOR}{\bm{u}}
\newcommand{\STATEPROPMAT}{_{\robotf}^{\mapf}\bm{T}\left(^\mapf\theta_t\right)}
\newcommand{\INPUTSPACE}{\mathcal{J}}
\newcommand{\BODYVELSPACE}{\mathcal{B}}
\newcommand{\CENTRIFUGAL}{\psi}
\newcommand{\OPTIMCALIBTIME}{$t_{\text{opt}}$\xspace}
\newcommand{\robotstate}{\bm{q}}
\newcommand{\mapf}{\mathcal{G}} 
\newcommand{\robotf}{\mathcal{R}} 
\newcommand{\learnedslip}{\bm{\Xi}}
\newcommand{\blrweights}{\bm{\gamma}}
\newcommand{\blrinputs}{^{\robotf}\bm{x}}
\newcommand{\MRSME}{\epsilon}

\renewcommand{\sp}[1]{\textcolor{orange}{#1}}
\newcommand{\vecbold}[1]{\bm{#1}}
\newcommand{\todo}[1]{\textcolor{blue}{TODO: #1}}
\newcommand{\transpo}[1]{#1^{T}}
\newcommand{\bodyframe}[0]{{}^{B}}
\newcommand{\wheelframe}[0]{{}^{W}}
\newcommand{\FIGCOMMENTS}[1]{\textcolor{red}{FIGCOMMENTS: #1}}
\newcommand{\mbs}[1]{\SI{#1}{\meter \per \second}}
\newcommand{\mbss}[1]{\SI{#1}{\meter \per \square\second}}

\newcommand{\TEXTCOMMENTS}[1]{\textcolor{GREEJ}{TEXTCOMMENTS: #1}}
\receiveddate{XX Month, XXXX}
\reviseddate{XX Month, XXXX}
\accepteddate{XX Month, XXXX}
\publisheddate{XX Month, XXXX}
\currentdate{XX Month, XXXX}
\doiinfo{XXXX.2022.1234567}

\markboth{}{Author {et al.}}

\title{DRIVE Through the Unpredictability: From a Protocol Investigating Slip to a Metric  Estimating Command Uncertainty}

\author{Nicolas Samson \authorrefmark{1}, William Larrivée-Hardy\authorrefmark{1},\\ William Dubois\authorrefmark{1}, Élie Roy-Brouard\authorrefmark{1}, Edith Brotherton\authorrefmark{1}, Dominic Baril\authorrefmark{1}, Julien Lépine\authorrefmark{2}, François Pomerleau\authorrefmark{1}, Senior Member IEEE}
\affil{Department of  Computer Science and Software Engineering, Université Laval, Québec, Canada}
\affil{Department of Operations and Decision Systems, Université Laval, Québec, Canada}
\corresp{Corresponding author: Nicolas Samson (email: nicolas.samson@norlab.ulaval.ca).}
\authornote{*This research was supported by the Natural Sciences and Engineering Research Council of Canada (NSERC) through the BESC M. }

\begin{abstract}
Off-road autonomous navigation is a challenging task as it is mainly dependent on the
accuracy of the motion model. 
Motion model performances are limited by their ability to predict the interaction between the terrain and the \acp{UGV}, which an onboard sensor can not directly measure. 
In this work, we propose using the \ac{DRIVE} protocol to standardize the collection data for system identification and characterization of the slip state space. 
We validated this protocol by acquiring a dataset with two platforms (from \SI{75}{kg} to \SI{470}{kg}) on six terrains (i.e., asphalt, grass, gravel, ice, mud, sand) for a total of \SI{4.9}{\hour} and \SI{14.7}{\kilo \meter}. 
Using this data, we evaluate the \ac{DRIVE} protocol's ability to explore the velocity command space and identify the reachable velocities for terrain-robot interactions.
We investigated the transfer function between the command velocity space and the resulting steady-state slip for a \acp{SSMR}.  
An unpredictability metric is proposed to estimate command uncertainty and help assess risk likelihood and severity in deployment.
Finally, we share our lessons learned on running system identification on large \ac{UGV} to help the community.

\end{abstract}



\maketitle

\acresetall 

\section{\MakeUppercase{Introduction}}

\IEEEPARstart{M}{otion} models are essential to multiple key tasks of autonomous navigation such as localization~\citep{Dumbgen2023}, path planning~\citep{Takemura2021} and path following~\citep{Brunke2022}.
For off-road navigation, the accuracy of the motion models is limited by their ability to model the terrain-robot interaction, which is not directly measurable by exteroceptive or proprioceptive sensors~\cite {Teji2023}.
Poor vehicle-terrain characterization will lead to significant modeling errors, potentially causing system failure~\citep{Seegmiller2016}. 
Furthermore, system identification of off-road \ac{UGV} is complex due to various factors that modify the traction between the vehicle locomotion mechanism and the terrain \citep{LOPEZ2021}.
These factors are called  \emph{factors of slip} for short and can be divided into two groups: internal and external. 
The internal factors are linked to the driving behavior, the vehicle's properties and their variation in time. 
The external factors are linked to the interaction between the \ac{UGV} and its environment, such as terrain steepness, roughness, and ground hardness.
The factors of slip must be well identified when deploying a \ac{UGV} on a given terrain as they have an important impact on the state variables of the \ac{UGV} and its 3D navigation dynamics.

The identification of these factors of slip requires deploying the \ac{UGV} in its operational environment and manually driving it for an extended period to gather a dataset that explores this high dimensional system~\citep{Williams2018}.
This exploration is usually based on two control variables in the body frame: the longitudinal and angular speeds.
This identification process remains challenging as the methodology of manual driving is not standardized, which creates a wide variety of motion complexity in the training and validation datasets.
The absence of a standard for data gathering strategy, combined with the information limitation of system identification of the terrain-\ac{UGV} interaction, reduces the reproducibility of the results.
Additionally, standardizing this process could help engineers ensure that their systems comply with the ISO 34502:2022(E) standard on autonomous navigation.\footnote[2]{“ISO 34502:2022(E): Road vehicles — Test scenarios for automated driving systems — Scenario-based safety evaluation framework”, 2022}

\begin{figure}[t!]
    \centering
    \includegraphics[width=0.48\textwidth]{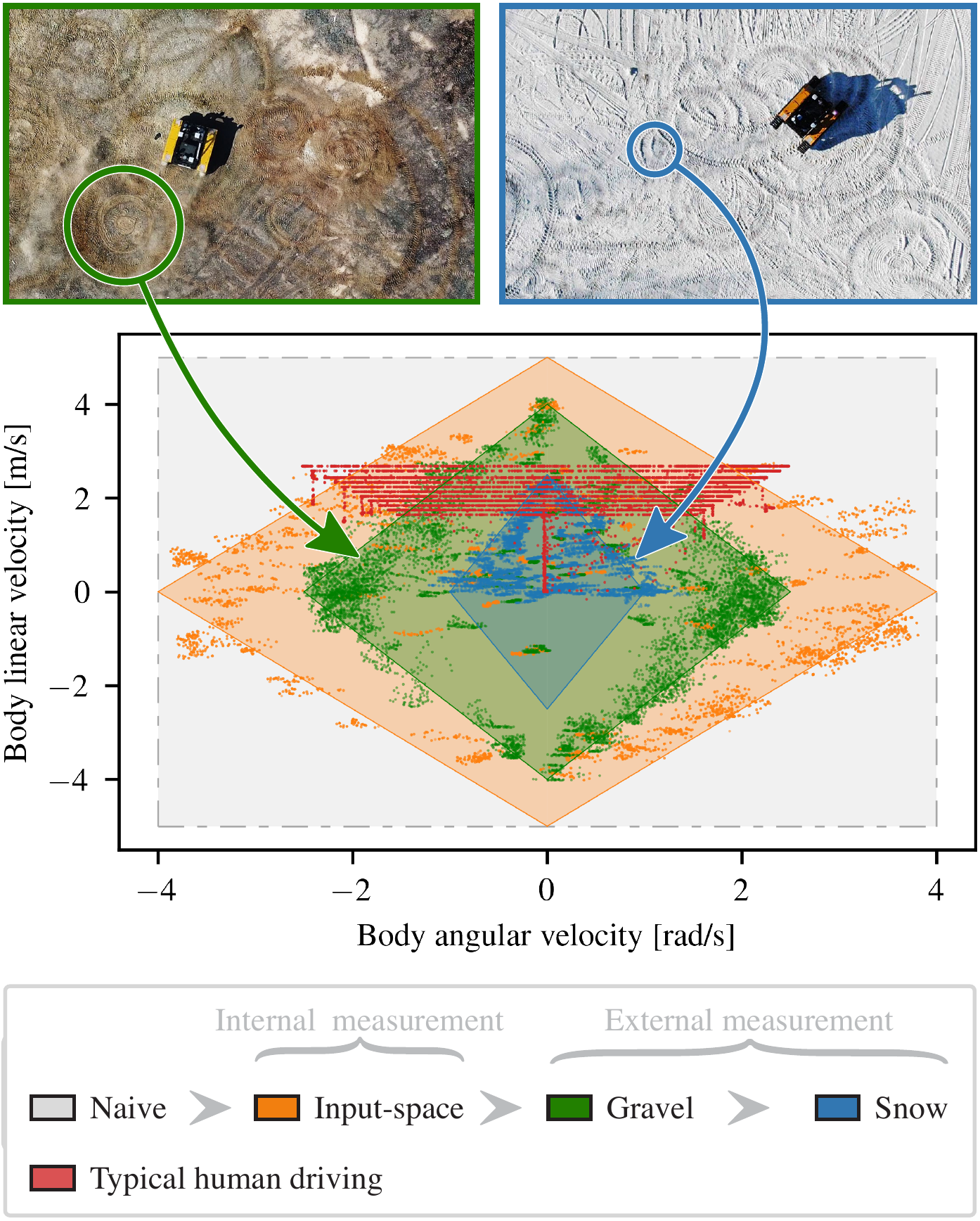}
    \vspace{-0.1in}
    \caption{
            Vehicle and terrain characterization done through~\DOUGHNUTCALIB.
            The manufacturer-defined Naive input-space region is drawn in gray.
            The vehicle's true input space, characterized through internal measurements, is shown in orange.
            Typical human driving is shown in red.
            The resulting body velocities are represented in green for gravel and blue for snow.
            }
	\label{fig:input-space}
\end{figure}

To address this issue, \citet{Baril2024} has proposed the protocol called ~\emph{\acf{DRIVE}}, which facilitates and standardizes the terrain-robot system identification as illustrated in~\autoref{fig:input-space}.
The first step of \ac{DRIVE} is to identify the true vehicle's input space, differing from the manufacturer's specifications.
The second step is to determine acceleration and speed limits to prevent damage to the vehicle.
The last step is to automatically send commands to the~\ac{UGV} to cover the entire true input space.
This differs from the common manual driving approach, which tends to cover only a small proportion of the input space, as shown by the red dots representing our previous work~\citep{Baril2020}.

This paper is an extension of \citet{Baril2024}'s work. 
The \ac{DRIVE} protocol has been deployed multiple additional times for a total of \SI{13.0}{\kilo \meter}  with a Clearpath Robotic Warthog on five terrains (i.e., sand, ice, asphalt, grass, and gravel) and  \SI{1.7}{\kilo \meter} with a Clearpath Robotic Husky on three different terrains (i.e., grass, asphalt, and mud).
This large dataset is the basis of the four contributions of this paper: 
\begin{itemize}
    \item the validation that sampling period of \ac{DRIVE} enables the characterization of the slip state space in steady state. 
    \item the analysis of the system identification over six terrains with two robots.
    \item a novel metric related to the \emph{unpredictability} of autonomous navigation.
    \item lessons learned on running system identifications on large \ac{UGV}.
\end{itemize}  

In this paper, \autoref{sec:rw} presents a review of the dataset-gathering protocol, followed by a review of the factors of slip for \ac{UGV} and how they have been integrated to different motion models. 
\autoref{sec:methodology} defines the vehicle motion state space and constraints for a typical \ac{SSMR}, explains the impact of terrain on commands, presents the \ac{DRIVE} protocol, and presents the formulation of the slip model and the unpredictability metric. 
\autoref{sec:experimental_setup} summarizes the deployments realized and presents the hardware used for the experiments. 
\autoref{sec:results} contains the results, which include the analysis of the \ac{DRIVE} protocol's ability to explore the steady-state velocities, the transfer functions between the command space and the slip for different terrains, and the unpredictability metric results that can be used to quantify the risk associated with kinetic energy for any robot motion on any terrain. 

\section{\MakeUppercase{Related Work}}
\label{sec:rw}
This section reviews existing dataset-gathering strategies and their limitations. 
The factors causing vehicle slippage, as identified in the motion modeling literature, are positioned against the general challenge of system identification. 
Finally, a few proposed metrics to quantify off-road dataset difficulty are presented to show explicitly the need for a metric that quantifies the robot-terrain interactions.

\subsection{\MakeUppercase{Dataset gathering protocol}}
Although many vehicle motion models require empirical training data, only a few dataset-gathering protocols have been published.
~\citet{Voser2010} have proposed to maintain a steady forward velocity while slowly increasing angular velocity, enabling the generation of a quasi-steady-state empirical dataset. 
~\citet{Wang2015} have proposed a similar approach with a varying commanded curvature radius to empirically identify the relation between angular velocity and \ac{SSMR} skid. 
Unfortunately, these simple approaches only cover a small subset of the total vehicle’s command space composed of longitudinal and angular speeds. 
These two approaches only sample small accelerations, which limits their application for learning the slip in a transient state. 
One can also find large, multi-modal datasets allowing to train and evaluate models for off-road and extreme driving ~\cite{Triest2022}. 
However, such datasets over-represent forward motion, are limited to a specific \ac{UGV} and would require new training data for any new vehicle configuration. 
Manual training data gathering guidelines have been proposed by  ~\citet{Williams2018}, asking the driver to vary his driving style. 
However, these remain time-consuming and subject to input space coverage bias. 
In ~\citet{Baril2024}, it is demonstrated that training a motion model with the \ac{DRIVE} protocol allows increased motion prediction performance and fast training dataset gathering.
This paper also demonstrated that covering the whole command space reduces the motion modeling error in translation and rotation.  

Extending the work of \citet{Baril2024}, we present here an updated version of the \ac{DRIVE} protocol.
The two major differences with the updated protocol version are additional information on the sampling space definition to avoid slip estimation error and an evaluation of the protocol's ability to sample the possible steady-state velocities for a given robot-terrain interaction. 
In this paper, steady state refers to when the robot velocities are constant through time. 
 
\subsection{\MakeUppercase{Factors of slip for uncrewed ground vehicle navigating off-road }}
\begin{figure}[h!]
    \centering
    \includegraphics[width=0.5\textwidth]{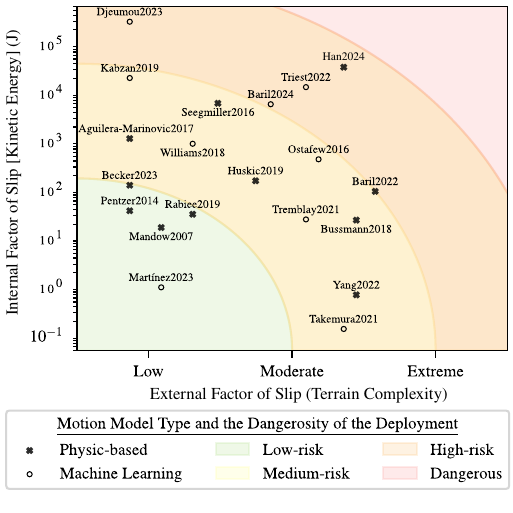}
    \caption{Qualitative mapping of the related work based on the factors of slip respectively represented by the maximum kinetic energy of the vehicles used in the experiment and the most complex terrain used in the experiment. 
    The type of motion model developed by the authors is also presented using different markers. 
    The area created by the internal and external factors of slip is also divided based on the risk associated with \ac{UGV} deployment in these conditions.
    One should note the logarithm scale used for the \textit{\textit{y}-axis.}}
    \label{fig:RW_qualitative}
\end{figure}

The factors causing the slippage of a \ac{UGV} can be divided into two groups: internal and external.
Internal factors of slip include factors that are caused by the vehicle properties or a command vector~$\INPUTVECTOR$. 
For example, commanding the \ac{UGV} to drift will generate more slip than turning slowly without skidding.  
External factors of slip include factors caused by the environment, such as terrain properties. 
To pose the robot-terrain interaction identification problem, we provide a comprehensive list for both groups as observed in previous~\ac{UGV} field deployments from different authors. 
In~\autoref{fig:RW_qualitative}, these deployments are mapped based on the experiment description made in their published article, with their external and internal factors of slip level represented on the \textit{x}- and \textit{y}-axes respectively.
The information on the factors of slip varies from paper to paper. 
Consequently, the internal factors of slip are approximated by the maximum kinetic energy of the heaviest \ac{UGV} used in the experiment. 
This approximation gives the order of energy engaged in the system.
Hence, this also estimates the order of the forces that must be applied to change the state of the \ac{UGV}.
Another benefit of approximating the internal factor of slip with the maximum kinetic energy is that it only depends on two commonly given parameters: the mass and the square of the vehicle's maximum speed.
So it is easy to compute for almost any deployment.

Again referring to \autoref{fig:RW_qualitative}, the external factors of slip are approximated by the most complex terrain tested in each article.
The terrain complexity has been \emph{qualitatively} ordered based on the information available about the terrain steepness, roughness, hardness, and friction coefficient. 
The terrain complexity goes from flat asphalt to soft soil such as sand and deep snow with steep slopes~\cite{Yang2022, Baril2022}.  
When not available, terrain information was inferred based on the visual information present in the article.

\textbf{Internal factors of slip}, which are related to the vehicle and how it is commanded, have been studied by many researchers.   
\citet{Rabiee2019}  have quantified the accuracy improvement when including the impact of vehicle inertia in the motion model. 
\citet{Kabzan2019} have shown that high-speed driving on a racetrack with an autonomous race car leads to tire saturation, causing slip, which they model with \acp{GP}.
\citet{Djeumou2023} combined this high-speed driving with aggressive motions, pushing the tires past the saturation point.
In their work, they successfully performed autonomous drifting with a full-size car by modeling tire forces with a neural network. The car weight (\SI{1450}{kg}) and top speed (\SI{13.86}{m/s}) make it the motion model validated with the highest kinetic energy.
\citet{Becker2023} also used tire force models to improve autonomous racing performance for a 1:10 scale race car.
Aggressive driving was also performed on a dirt racetrack by \citet{Williams2018}, who modeled slip with a neural network. However, both authors used a scaled-down car version, which results in lower kinetic energy.
In their work, they have shown that aggressive motion on dirt roads leads to a change in the vehicle orientation, which affects the resulting body velocity. 
The steering geometry also has an impact on the internal factors of slip.
For example, Ackermann steering geometries are designed to minimize wheel skidding, while~\acp{SSMR} require skidding for angular motion~\citep{Mandow2007}. 
\citet{Pentzer2014} have shown that the inherent skidding and slipping of \ac{SSMR}s increased with the speed of the \ac{SSMR}.  

\textbf{External factors of slip} are related to the surrounding environment and the vehicle's interaction with that environment. 
A small change of terrain properties between the ground at calibration and the ground at deployments can lead to a divergent motion model if the parameters are not updated continuously~\cite{Bussmann2018}. 
\textbf{Terrain steepness} has multiple impacts on the vehicle motion. 
\citet{Seegmiller2016} have shown that the vehicle pitch and roll orientation affect motion to a scale depending on the terrain properties.
\citet{Ostafew2016} have observed a similar phenomenon when driving a \ac{SSMR} over gravel piles.
\citet{Takemura2021} deployed a lunar rover on loose terrain with an inclination leading to ground deformation and terrain slip.
As part of their large off-road driving dataset, \citet{Triest2022} deployed an \ac{ATV} weighing approximately \SI{730}{kg} in sloped terrains and benchmarked learning approaches to account for the resulting slip. 
Steep terrain also limits the ability of \ac{SSMR} to turn around itself because its inherent skidding combined with the slopes results in the \ac{SSMR} skidding towards the bottom of the slopes instead of skidding around itself~\citep{Martinez2023}.  

\textbf{Terrain roughness} can affect vehicle motion by reducing the contact between the ground and the wheels. A rougher terrain leads to a higher number of contact losses and, therefore, a more complex motion. 
\citet{Yang2022} have explained that a wrong estimation of the contact area between the deformable soil and the wheels of a \ac{UGV} can lead to significant motion prediction error and the failure of the dynamic-motion-model simulation. 
Similarly, \citet{Tremblay2021} deployed a small-size~\ac{SSMR} in boreal forest trails cluttered with fallen branches and roots that modify the contact between the ground and the wheels.
In their work, they learn the slip based on multiple sensor modalities. 
In parallel, \citet{Aguilera-Marinovic2017} developed another general physic-based motion model approach that can estimate the impact of the loss of contact between the ground and the wheels on the movement of a mobile manipulator installed on a \ac{SSMR}. 

\textbf{Ground hardness and friction} are also important external factors of slip. 
\citet{Fiset2021} have studied the energy-loss and ground bulldozing effect for a~\ac{SSMR} turning in loose sand. 
In previous work, we have shown that deep snow navigation can lead to complete vehicle immobilization, which would represent an extreme case of motion distortion~\citep{Baril2022}.
As for terrain friction, \citet{Huskic2019} have shown that the same set of parameters was sufficient to conduct high-speed path following on relatively uniform terrain, namely asphalt, light grass, and gravel.
However, in previous work, we have shown that navigation on surfaced ice increased the slip compared to other hard ground~\citep{Baril2024}. 
In this case, we observed that the low friction force leads to longer transient behavior and a high impact of vehicle inertia.
Even advanced slip learning-based models failed to predict motion accurately in this terrain \citep{Baril2024}.
As we can see, many factors influence the behavior of a robot and the current literature is evasive in describing the tested environment.
Therefore, there is a need for a more standard protocol to quantify the robot-terrain interaction, thus improving the quality of comparison between scientific outcomes.

\subsection{\MakeUppercase{Comparing motion models}}
As shown in \autoref{fig:RW_qualitative}, motion models are often tested in a limited variety of factors of slip, which makes it hard to compare the performances between motion models. 
For example, \citet{Yang2022} have a motion model that performed well in adverse external factors of slip like non-cohesive sand and slopes but was only tested for low internal factors of slip. 
\citet{Djeumou2023} did the opposite by validating its motion model only in the presence of high internal and low external factors of slip. 
A factor of slip metric agnostic to the causes could help reduce the different factors of slip in one dimension.

Few metrics have been published to quantify the internal or external factors of slips.
\citet{Triest2022} have estimated the difficulties of their dataset using the median variation of height per second, assuming that steepness and unevenness of the trajectories are key factors of a dataset's difficulty. 
\citet{hanWICRA2024} have proposed a more generic approach to define metrics using the distance between the trajectory states and their respective distribution in a reference dataset. 
The case studied in the results used the trajectories accelerations and training dataset containing vertical accelerations of $9.8\pm$~\SI{2.7}{\meter \per \square \second} to score the aggressiveness of each trajectory. 
The aggressiveness of new trajectories is relative to the reference dataset.
Thus, the validity of the aggressiveness metric highly depends on the dataset-gathering protocol used to collect the dataset, which highlights the importance of a standard dataset-gathering protocol.
The metric accuracy also depends on the factors of slip similarity between the reference dataset and the new trajectories evaluated.   
\citet{SamsonWICRA2024} have proposed another approach that used the error of an ideal motion model to estimate the difficulties of a dataset.
The underlying hypothesis is that the more complex the trajectories of a dataset are, the greater the error of an ideal motion model that assumes no slip will be. 
One of the limitations of this technique is that it depends on the realism of the command sent to the motors. 
In other words, unachievable commands increase the difficulty of the dataset. 
Another shortcoming is that the technique combines the translation and angular error into one scalar, and as both do not share the same physical unit, there can be a scaling issue. 
To overcome these challenges, this paper proposes the unpredictability metric, which evaluates the realism of a command given a robot-terrain interaction. 
This metric has the advantage of being agnostic to the factors of slip but still measures their impact on the vehicle motions.  
It can also be used to estimate the likelihood of a given command, which is useful to assess a deployment risk. 

The mapping in \autoref{fig:RW_qualitative} also highlights a remaining research challenge: \emph{identification of terrain-robot interaction in high-risk deployments}. 
To the authors' best knowledge, only \citet{Han2024,Triest2022,Djeumou2023} have pushed the identification of terrain-robot interaction in high-risk deployments. 
\citet{Triest2022} gathered a dataset with a Yamaha Viking navigating on a forest track with a median height variation of \SI{0.23}{m/s} and a vehicle speed of up to \SI{6}{m/s}. 
Similarly, \citet{Han2024} deployed an autonomous \ac{ATV} that successfully navigated at a speed of up to \SI{10}{m/s} on different complex terrains such as v-ditches, dried rivers, slopes of 20 degrees, and craters. 
However, both of these deployments have a respective maximum kinetic energy that represents only \SI{1.5}{\%} and \SI{0.11}{\%} of the kinetic energy used to validate the motion model of \citet{Djeumou2023}.   
In addition, some terrains remain out of reach. 
For example, \citet{Baril2022} deployed a \ac{SSMR} in snow and observed that deep snow is a highly complex terrain to navigate because the \ac{SSMR} tends to sink in it. 
The sinkage of the vehicle reduces the viable angular motion. 
If the turning restrictions are not respected, the \ac{SSMR} can sink to the point where it is impossible to recover without human intervention. 
As risk evaluation tends to be subjective, finding metrics that allow comparison of risk levels would give an independent assessment of different field deployments involving autonomous navigation.

\section{\MakeUppercase{Methodology}}
\label{sec:methodology}

In this section, we provide details on the definition of the vehicle's motion state space, the constraints that apply to these states, and how the perturbations affect the state space.
The \ac{DRIVE} protocol is described, followed by a section explaining the process of computing the relationship between the command space and the different slip dimensions.  
Finally, the unpredictability metric highlighting how well a robot can anticipate the future is explained.

\subsection{\MakeUppercase{Definition of the search space and its constraints}}
To gather a dataset that quantifies the consequence of factors of slip on the \ac{UGV} state, it is first necessary to define the \ac{UGV} state vector $\vecbold{x}$ and the constraints that apply to these states.
This state vector can be directly observed or estimated from sensory inputs that we express as the observation vector $\vecbold{z}$.
Converting observations $\vecbold{z}$ to state $\vecbold{x}$ is a field of research on its own and to keep this paper contained, we will assume that we can access directly the state vector $\vecbold{x}$ and its noise is reasonable.
The first six variables of $\vecbold{x}$ describe the pose $\vecbold{p}$ of the \ac{UGV}, such that $\vecbold{p} = \left[ \vecbold{t}, \vecbold{\theta} \right]^T$, where  $\vecbold{t} \in \mathbb{R}^3$ is the translation states and $\vecbold{\theta} \in \mathfrak{so}(3)$ is the attitude states.
The following 12 states describe the velocity $\vecbold{\dot{p}} \in \mathbb{R}^6$ and the acceleration  $\vecbold{\ddot{p}} \in \mathbb{R}^6$ of a rigid body.
More specifically for \acp{UGV}, we need to define contact points with the ground.
\acp{UGV} will have $i$ contact points representing wheels or tracks actuated by a motor producing a rotational velocity $\omega_i$ and rotational acceleration $\dot{\omega_{i}}$, which can be concatenated as $\vecbold{\omega} \in \mathbb{R}^i$ and $\vecbold{\dot{\omega}} \in \mathbb{R}^i$.
Finally, vehicles with active steering or articulations have $j$ number of joints with an angle $\phi_{j}$ and a rotational velocity $\dot{\phi_{j}}$, which can also be concatenated as $\vecbold{\phi} \in \mathbb{R}^j$ and $\vecbold{\dot{\phi}} \in \mathbb{R}^j$.
Thus, we end up with $18+2i+2j$ state variables to define motion in space for any \ac{UGV}.
On the other side, the command vector $\vecbold{u}$ will only include the wheel velocity $\vecbold{\omega} \in \mathbb{R}^i$ and the joint angle $\vecbold{\phi} \in \mathbb{R}^j$.
For convenience, the input $\vecbold{u}$ is often defined in the body frame $B$ of the robot instead of for each actuator frame of reference.
Using reverse kinematic through a \emph{motion model} $\text{m}(\cdot)$, we can map commands for linear and rotational velocities $\vecbold{\dot{p}}$ expressed in the body frames to each actuator and reverse.
This motion model needs to make assumptions on the \ac{UGV}-terrain interactions, which is the main goal of our proposed protocol.
No matter how the command vector $\vecbold{u}$ is expressed, this configuration leads to fewer inputs than excited states, such that $\text{dim}(\vecbold{x}) > \text{dim}(\vecbold{u})$, with 
\begin{equation}
\label{eq:states}
\vecbold{x} = 
\begin{bmatrix}
\vecbold{p} \\
\vecbold{\dot{p}} \\
\vecbold{\ddot{p}} \\
\vecbold{\omega} \\
\vecbold{\dot{\omega}} \\
\vecbold{\phi} \\
\vecbold{\dot{\phi}}
\end{bmatrix}
\; \text{and} \;
\vecbold{u} = 
\begin{bmatrix}
\vecbold{\omega} \\
\vecbold{\phi} \\
\end{bmatrix} 
\longleftrightarrow
\vecbold{u} = 
\begin{bmatrix}
\vecbold{\dot{p}}
\end{bmatrix} 
.
\end{equation}

In addition, each dimension of the search space is affected by a whole spectrum of factors of slip, which implies surveying multiple types of ground. 
The number of combinations of factors of slip grows exponentially with the dimensions of the states to observe, which need to be repeated on a variety of terrains.
This procedure is challenging at best when it comes to deploying \acp{UGV} in the field for each recording.
Therefore, the \ac{DRIVE} protocol aims to automatize dataset gathering and slip identification for a given \ac{UGV} by relying on approximations to reach a tractable sampling problem. 
The underlying assumptions are that the designed
area for slip identification is:
\begin{enumerate}
    \item \emph{reasonably flat}, such that a pose in two dimensions approximates the vehicle slip.
    Moreover, the effect of gravity on the motion of the \ac{UGV} can be neglected thus focusing only on the \ac{UGV}-terrain interaction.
    This assumption reduces the number of states used to track the rigid body in space from 18 to nine.
    \item \emph{reasonably uniform} in terms of soil composition and that the \ac{UGV}'s motion on that soil does not alter too much its factor of slip. 
    A case of alteration would be an \ac{UGV} in deep snow generating a layer of ice by repeatably driving at the same location.
\end{enumerate}
We use the term \emph{reasonable} to highlight that these recordings on different terrain should be collected in representative areas, which cannot be perfectly controlled.
Finding terrain large enough to conduct a recording requires balancing flatness and uniformity against safety.
It is assumed that the resulting slip includes noise and represents the average slip for a given terrain.

\subsection{\MakeUppercase{Search space for a typical \ac{SSMR}}}

\begin{figure}[htbp]
    \centering
    \includegraphics[width=\linewidth]{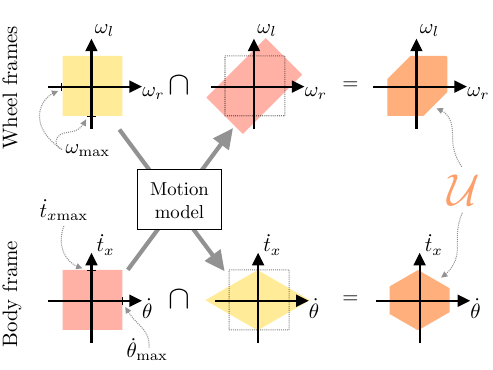} 
    \caption{Visualization of the command space $\boldsymbol{\mathcal{U}}$ reduction caused by different thresholds.
    On the first row, the command space in the wheel frames is shown as a yellow square bounded by $\boldsymbol{\omega_\text{max}}$, which results in a tilted diamond after being intersected with the constraints set in the body frame.
    On the second row, the command space in the body frame is shown as a red square bounded by $\boldsymbol{\dot{t}_{x\text{max}}}$ and $\boldsymbol{\dot{\theta}_{\text{max}}}$, which results in a hexagon after being intersected with the constraints set in the wheel frames.} 
    \label{fig:commandspace}
\end{figure}

We can reduce even further the search space when it comes to a typical \acf{SSMR} that has only one motor controlling the right side $r$ and one other controlling the left side $l$.
As there is no joint angle, we can simplify the command vector to $\vecbold{u_\omega} = \begin{bmatrix} \omega_l & \omega_r \end{bmatrix}^T$.
In practice, these commanded velocities expressed in their respected wheel frame $W$ are bounded by a low-level motor controller limiting the domain to $\vecbold{u_\omega} \in [-\omega_\text{max}, \omega_\text{max}]^2$.
One should note that typical low-level motor controllers follow a maximum wheel velocity ramp, which is equivalent to a maximum wheel acceleration ($\dot{\omega}_\text{max}$), but we neglect this parameter as it is challenging to reverse engineering if not openly provided.
The manufacturer of the \ac{UGV} also provides maximum linear velocity $\dot{t}_{x\text{max}}$ and rotational velocity $\dot{\theta}_{\text{max}}$, but this time in the \ac{UGV}'s body frame $B$.
As these thresholds live in different reference frames, we need an invertible motion model $\text{m}(\cdot)$ allowing us to express these search boundaries in all frames.
\autoref{fig:commandspace} illustrates how these thresholds interact through the motion model, leading to a bounded search space of possible commands often smaller than what a user might expect when only controlling in the body frame $B$.
Another typical threshold defined by the user to avoid premature wear of the vehicle is a maximum linear acceleration $\ddot{t}_{x\text{max}}$ in the body frame $B$.
This threshold doesn't bind the command space $\mathcal{U}$, but rather the distance between a sequence of two consecutive commands $\vecbold{u}$ in that space.
Finally, the maximal lateral speed $\dot{t}_{y\text{max}}$ and maximal lateral acceleration $\ddot{t}_{y\text{max}}$ is not defined by the user for a \ac{SSMR} because of the non-holonomic nature of the skid-steer. 

As mentioned before, to transform the wheel speed constraints to the body $B$ or the body speed constraint into the wheel frame $W$, it is essential to have a motion model $\text{m}(\cdot)$ to change reference frames.
The most common motion model used by robot manufacturer  for \ac{SSMR} is the \ac{IDD} motion model \citep{Baril2020}, which approximate the function $\text{m}(\cdot)$ using the matrix $\vecbold{M}$ such that
\begin{equation}
\label{eq:motionmodel}
\underbrace{
    \begin{bmatrix}
    \dot{t}_x \\[0.5em]
    0 \\[0.5em]
    \dot{\theta}
    \end{bmatrix}
}_{{}^{B}\vecbold{u_{\dot{p}}}}
    = 
\underbrace{r
    \begin{bmatrix}
         \phantom{-}\frac{1}{2} & \frac{1}{2}  \\[0.5em]
         0 & 0 \\[0.5em]
         -\frac{1}{b} & \frac{1}{b} 
    \end{bmatrix}}_{\vecbold{M}}
    \underbrace{
    \vphantom{\begin{bmatrix} \dot{t}_x \\[0.5em] 0 \\[0.5em] \dot{\theta} \end{bmatrix}}
    \begin{bmatrix}
    \omega_l \\[0.5em]
    \omega_r
    \end{bmatrix}
    }_{{}^{W}\vecbold{u_\omega}},
\end{equation}
where $b$ is the base width, $r$ the wheel radius.
Note that $\vecbold{M}$ is invertible, so we can express the command vector in the body frame ${}^{B}\vecbold{u}$ or in each wheel frame ${}^{W}\vecbold{u}$ depending of the needs. 
Moreover, the combination of the 2D assumption with the configuration of the \ac{SSMR} (i.e., $i=2$ and $j=0$) leads to a reduced dimension of the state vector with $\text{dim}(\vecbold{x}) = 11$, which is still larger than the command vector dimension which is two.

\subsection{\MakeUppercase{Impact of terrain on reachable commands}}
\label{sec:theory:constraints}
\begin{figure}[htbp]
    \centering
    \includegraphics[width=\linewidth]{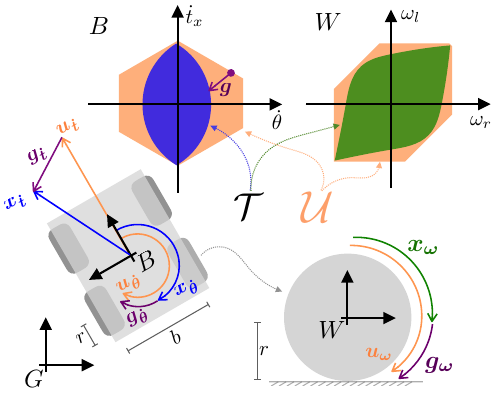}
    \caption{Representation of the robot navigating in 2D on a terrain that induces different slip $\boldsymbol{g}$ between the commanded $\vecbold{u}$ and realized motion $\boldsymbol{x}$ in the body frame $\boldsymbol{B}$ and the wheel frames $\boldsymbol{W}$.
    The original command space $\mathcal{U}$ is reduced by different factors of slip resulting in a smaller measured controlled space $\mathcal{T}$ represented in blue in the body frame $B$ and in green in the wheel frame $W$.}
    \label{fig:mathematical_definition}
\end{figure}

Sending commands in the body frame using an \ac{IDD} model relies on two main assumptions: 1) a step command in velocity is applied punctually (i.e., neglecting the inertia of a vehicle), and 2) all the energy is used for motion (i.e., there is no lost in friction).
But in reality, the \ac{UGV}-terrain interaction will induce slip on the commanded velocities.
Thus, once the command space of the vehicle ${}^{B}\mathcal{U}$ is defined as the intersection of the user limits and physical limits, it is possible to deploy the vehicle on the field and calculate the effect of this factor of slip. 
The terrain factors of slip impose additional bounds on the command search space by defining the traction between the tire and the terrain.  
This is illustrated at \autoref{fig:mathematical_definition} where the resulting command space of a vehicle on a specific terrain $\mathcal{T}$ (green and blue) is a subset of the command space $\mathcal{U}$ (orange). 
In this work, we neglect the vehicle dynamics by focusing our observations on the quasi-steady state of a command leading to the remaining error caused by the \ac{UGV}-terrain interaction.
Thus, the constraints imposed by the interaction of the wheels with the terrain will be referred to as the steady-state slip in the vehicle frame ${}^{B}\vecbold{g}$ and in the wheel frame as ${}^{W}\vecbold{g}$. 
In the wheel frames $W$, the steady-state slip $\vecbold{g}$ is defined as  
\begin{equation}
	 {}^{W}\vecbold{g} = \vecbold{u_{\omega}} - \vecbold{x_{\omega}},
	\label{eq:cmd}
\end{equation}
where $\vecbold{u_{\omega}}$ is the command of the left and the right wheel speed and $\vecbold{x_{\omega}}$ are the realized wheel speeds.

In the body frame $B$, the steady-state slip $\vecbold{g}$ is define as  
\begin{equation}
	 {}^{B}\vecbold{g} = \vecbold{u_{\dot{p}}} - \vecbold{x_{\dot{p}}},
	\label{eq:slip_body_def}
\end{equation}
where $\vecbold{u_{\dot{p}}}$ is the commanded velocities, $\vecbold{x_{\dot{p}}}$ is the realized velocities, both of which combine linear and rotational velocities. 

Both equations are visually represented in \autoref{fig:mathematical_definition}, where we also introduce a global reference frame $G$ where a robot can execute a given trajectory (i.e., a sequence of command $\vecbold{u}$ in time).
An extreme example of slip observed in the wheel frames $W$ is a ground offering too much friction (e.g., mud), which leads to a motor not having enough torque to reach a target velocity.
Equivalently, an extreme example of slip only observed in the body frame $B$ would be terrain with little friction (e.g., ice) leading to each wheel easily reaching its commanded velocity, with the center of mass of the \ac{UGV} not moving.

\subsection{\MakeUppercase{Data-driven robot input vector exploration (DRIVE)}}
\label{sec:protocol}
In the following section, the DRIVE protocol will be explained in detail for the system identification of a \ac{SSMR}, but the protocol can be generalized to other \acp{UGV}. 
In our use case, the only variable that can be used to explore the state space $\mathcal{S}$ is the wheel-speed target ($\omega_l$, $\omega_r$). 
To cover that search space, we rely on a sequence of \emph{random uniform sampling} of the~\ac{UGV} command space $\mathcal{U}$.
Although simple, this sampling method ensures a larger coverage of dynamic and transient vehicle behavior.
The system requirements to perform our protocol are as follows:
(i) a sub-servo system mapping body-level commands to wheel commands;
(ii) vehicle acceleration limits to reduce strain on vehicle components;
(iii) an accurate localization system, estimating the \ac{UGV} pose and velocity in a global frame; 
(iv) a safety operator to prevent the vehicle from leaving a predefined safe perimeter during the protocol.

The first step of~\DOUGHNUTCALIB is to identify user-defined limits on the system.
Therefore, one must identify the data processing pipeline that exists between a command sent, often in the body frame $B$, to the wheel movement of the \ac{SSMR}.
A typical architecture would have a high-level computer running data-intensive computation, a low-level computer converting commands in the body frame into separate motor commands, and multiple motor controllers ensuring velocity profiles at each wheel.
The acceleration and the velocity thresholds of each sub-system along that pipeline must be noted down to identify the most restrictive limits.
In the case of a symmetric \ac{SSMR}, one should expect the same threshold for all wheels and for both directions, leading to $\omega_\text{min} = -\omega_\text{max}$.
Then, by combining both the user-defined and physical limits, the input space~$\mathcal{U}$ is obtained in terms of body velocities and wheel velocity (i.e., recall \autoref{fig:mathematical_definition}). 
The second step of \ac{DRIVE} involves exploring the command space within the input limits with the vehicle on a given terrain. 
The strategy to explore the state space is to pick a random input vector~$\vecbold{u}$ defined as a two-dimensional uniform distribution parametrized by vehicle input limits such that $\INPUTVECTOR_{\omega} \sim \wheelframe\mathcal{U}$.
Thus, we can build a set $\mathcal{S}$ of sampling command where $\mathcal{S} = \left\{ \vecbold{u}_1, \vecbold{u}_1, \dots, \vecbold{u}_s \right\}$ with a total of $s$ sampled commands.
Our goal is then to apply each sampled input for a duration sufficiently long to gather steady-state motion data. 

\begin{figure}[htbp]
	\centering
	\includegraphics[width=\linewidth]{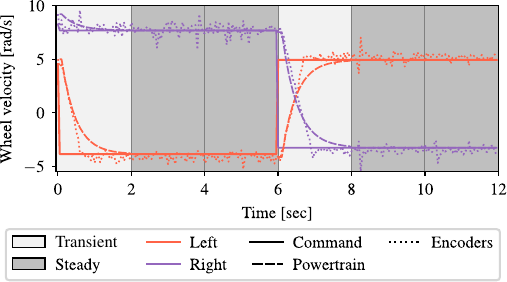}
	\caption{
            Sampling windows split into transient and steady-state sections for different representations of wheel velocities.
            Solid lines are the commands, dotted lines are from the encoder measurements, and dashed lines are a powertrain model~\citep{Baril2024} for both sides of a~\ac{SSMR} during two~\DOUGHNUTCALIB training intervals of six seconds each. 
            Each step consists of one transient-state window (in light gray) and two steady-state windows (in dark gray).
            Commands and measurements on the \textit{x}-axis are acquired at a rate of~\SI[detect-weight=true,mode=text]{20}{\hertz}.
            }
	\label{fig:transitory_steady_state}
\end{figure}

We define a two-second time window to stay compatible with the motion modeling literature such as defined in the seminal work on \ac{UGV} path following \citet{Williams2018}. 
Since the majority of~\ac{UGV} motion is quasi-steady state~\citep{Voser2010}, we define a training interval consisting of one transient training window and two steady training windows, lasting a total of six continuous seconds.
An example of two training intervals is shown in~\autoref{fig:transitory_steady_state}.
We keep both steady-state and transient-state windows to enable both steady-state and transient slip identification and evaluation.
\autoref{alg:drive} show in more detail what is happening during the recording.
Lastly, we release our protocol as an open-source package to facilitate applicability.\footnote{https://github.com/norlab-ulaval/DRIVE} 

\begin{algorithm}[htbp]
\caption{\ac{DRIVE} protocol algorithm.}\label{alg:drive}
\begin{algorithmic}

\STATE \hspace{-0.2cm}{\textbf{function} \textsc{DRIVE}}$({}^{W}\mathcal{U}, h_{calib},s)$
\STATE \hspace{-0.2cm}\textbf{Input: }${}^{W}\mathcal{U}$ - \text{Sampling space in the wheel frame}
\STATE \hspace{-0.2cm}\textbf{Input: } $h_{calib}$ - \text{Calibration step duration}
\STATE \hspace{-0.2cm}\textbf{Input: } $s$ - \text{Number of sampled commands}
\STATE \hspace{-0.2cm}\textbf{Output:} $\mathcal{T}$ - \text{Set of measured states}
\STATE \hspace{-0.2cm}\textbf{Output:} $\mathcal{S}$ - \text{Set of command samples}
\STATE $\mathcal{S} \gets \{\}$
\STATE $\mathcal{T} \gets \{\}$
\FOR{ $n=1:s $ } 
\STATE $ \textbf{Select randomly } {}^{W}\vecbold{u}_n \in {}^{W}\mathcal{U}  $
\STATE $h \gets 0.0$
\STATE $\mathcal{S} \gets \mathcal{S} \cup \{{}^{W}\vecbold{u}_n\}$
\STATE $\vecbold{X} \gets \{\}$
\WHILE{$h < h_{calib}$}
\STATE $ h \gets  \textsc{SendCommandToMotors}({}^{W}\vecbold{u}_n)$
\IF{\texttt{Manually Interrupted}}
\STATE $h \gets 0.0$
\STATE $\vecbold{X} \gets \{\}$
\STATE \texttt{Reorient the vehicle to complete the interrupted command in the direction of the center of the safe zone. Then resume.}
\ENDIF
\STATE $\vecbold{X} \gets \vecbold{X} \cup \{\textsc{StateEstimation()}\} $
\ENDWHILE
\STATE $\mathcal{T} \gets \mathcal{T} \cup \{{}^{W}\vecbold{X}\}$
\ENDFOR
\RETURN $\mathcal{T} \textit{, } \mathcal{S} $
\end{algorithmic}
\end{algorithm}

The number of commands to be completed depends on the objective for which \ac{DRIVE} is used.
In this paper, DRIVE is used for two goals: understanding the relationship between the command space $\mathcal{U}$ and the slip for a \ac{SSMR} navigating on a given terrain while investigating a metric estimating the predictability of a vehicle.
The minimal number of completed commands has been experimentally determined and is motivated in our lessons learned (\autoref{sec:lesson_learned:deploy_drive}).
After processing the recorded data, the \ac{DRIVE} protocol provides the subset of reachable speed on a given terrain as well as the relationship between the command and the three dimensions of slip.
Both of this information could be implemented in a controller to improve its accuracy and solve the path-following problem.

\subsection{\MakeUppercase{Steady-state slip investigation}}
\label{sec:theory:slip_analysis} 

As \citet{Baril2024} demonstrated how the \ac{DRIVE} protocol can be used to learn the transient states of the vehicle using a first-order model, this paper will focus mainly on the vehicle slip in steady-state. 
Four steps are used to investigate the impact of the different terrains on the command space $\mathcal{U}$:
1) perform a DRIVE on a given terrain; 
2) compute the steady-state slip for all the steps; 
3) compute a uniform discrete transfer function between slip and command space inputs using a Gaussian kernel smoother; 
4) repeat steps two and three for both slip dimensions (i.e., the body speed $\bodyframe{\vecbold{g}}$ and the wheel speed $\wheelframe\vecbold{g}$). 
The first and second steps have been described in detail in  \autoref{sec:protocol} and in \autoref{sec:theory:constraints}. 
Thus, this section focuses on how the sampled space $\mathcal{S}$ can be converted into uniform discrete transfer functions between the command space and the slip using a Gaussian kernel smoother. 
To approximate the slip functions over the command space, we start by defining the interpolation distance for each dimension of the command space $\mathcal{U}$. 
Using these distances, a uniform grid of $n \times m$ elements is created over $\mathcal{U}$.
Then, we use a 2D Gaussian kernel for each interpolated point $\vecbold{\bar{u}}$ in the grid to compute the average slip $\vecbold{\bar{g}}$ based on the measured slip $\vecbold{g}$ resulting from a sampled command $\vecbold{u}$.  
The distance between the sampled commands and the point of the grid is used together with the Gaussian kernel to compute the weight $k$ of each $x$ sample $i$ using
\begin{equation}
    k_i = \frac{1}{2\pi \sqrt{\left| \vecbold{\Sigma} \right| }}
    \text{exp}\left[ (\vecbold{u} - \vecbold{\bar{u}})^T \vecbold{\Sigma}^{-1} (\vecbold{u} - \vecbold{\bar{u}}) \right],
	\label{eq:gaussian_kernel}
\end{equation}
where $\vecbold{\Sigma}$ is diagonal with user defined variance.
This equation gives us the relative importance of the sample $i$ to compute the slip at the interpolated command $\vecbold{\bar{u}}$.

Then, we can compute the weighted average slip $\vecbold{\bar{g}}$ for all points in the grid using
\begin{equation}
    \vecbold{\bar{g}} = \sum_{i=0}^{s}{\frac{{k_i} \vecbold{g}_i}{\sum_{i}^{s}{k_i}}}.
    \label{eq:average_slip}
\end{equation}

\subsection{\MakeUppercase{Unpredictability metric}}
\label{sec:theory:energy_metric}

The goal of the \emph{unpredictability metric} $\rho$ is to quantify how much a system can anticipate the link between a command and the produced motion of a vehicle. 
\citet{SamsonWICRA2024} proposed to use the distribution formed by the norm of the error of the \ac{IDD} (i.e., the slip vector $\vecbold{g}$). 
However, this method is limited by the use of arbitrary weights to combine the translational (m/s) and rotational errors (rad/s).
The kinetic energy-based readily combines these units through an inertia matrix $\vecbold{I}$.
We can compute this inertia matrix $\vecbold{I}$ by approximating the robot as a rectangular prism with uniform density, leading to
\begin{equation}
\label{eq:inertia}
    \vecbold{I} = \frac{m}{12} 
    \begin{bmatrix}
        (b^2 + c^2) & 0  & 0\\
         0 &  (d^2 + c^2)  & 0\\
         0 & 0 &  (d^2 + b^2)  
    \end{bmatrix},
\end{equation}
$b$, $d$, and $c$ are the vehicle base width, depth, and height respectively.

The total kinetic energy $K \in \mathbb{R}_{\leq 0}$ of a \ac{UGV} is the sum of the rotational kinetic energy $K_{\dot{\theta}}$ and the translational kinetic energy $K_{\dot{t}}$ defined as
\begin{equation}
\label{eq:Ek}
\begin{split}
    K_{\dot{\theta}} &= 
    \frac{1}{2}  
    \left( \transpo{\vecbold{\dot{\theta}}} \vecbold{I} \vecbold{\dot{\theta}} 
    \right) \quad \text{and}
\\
    K_{\dot{t}} &= 
    \frac{m}{2}  
    \left(
    \transpo{\vecbold{{\dot{t}}}} \vecbold{\dot{t}} 
    \right),
\end{split}
\end{equation}
where $m$ is the mass of the vehicle in kg.

Comparing the kinetic energy commanded to the one measured would only tell us if the magnitudes are similar.
Thus, we added the terms $\{\alpha, \beta\} \in [0, 1]^2$ that will penalize any misalignment of the translational and rotational velocity vectors, such that
\begin{equation}
\label{eq:orientation_penalities}
\begin{split}
    \alpha &= \frac{1}{2} 
    \left( \vphantom{\frac{1}{2}}
    \vecbold{\hat{u}_{\dot{t}}} \cdot \vecbold{\hat{x}_{\dot{t}}} + 1 
    \right)
    \quad \text{and}
\\
    \beta &= 
    \frac{1}{2}  
    \left( \vphantom{\frac{1}{2}}
    \vecbold{\hat{u}_{\dot{\theta}}} \cdot \vecbold{\hat{x}_{\dot{\theta}}} + 1 
    \right),
\end{split}
\end{equation}
where $\hat{\square}$ is a normed vector.
Then, we can compute the commanded total kinetic energy $K_{u} = K_{u\dot{t}} + K_{u\dot{\theta}}$ and its resulting weighted total energy kinetic $K_{x} = \alpha K_{x\dot{t}} + \beta K_{x\dot{\theta}}$.

The unpredictability metric $\rho$ uses a ratio of the measured kinetic energy $K_x$  over the commanded kinetic energy $K_u$ to quantify the energy loss in the actuation. 
Finally, we can avoid division by zero using the $\text{arctan2}(\cdot)$ and rescalling factors to ensure that $\rho \in [0, 1]$ using 
\begin{equation}
\label{eq:tm}
    \rho(\vecbold{u_{\dot{p}}}, \vecbold{x_{\dot{p}}}) = \frac{4}{\pi}  \abs{\text{arctan2}{\left(
    K_u, K_x
    \right)} -\frac{\pi}{4}}.
\end{equation}

The first advantage of this metric is its proportionality to the energy loss in the terrain.
When the ground absorbs a negligible amount of energy, the ratio between the kinematic energy computed from the command $K_{u}$ and the kinetic energy calculated from the measurements $K_{x}$ should tend towards one in steady state. 
Therefore, $\rho$ will tend towards one when the terrain completely immobilizes the vehicle, as $K_{x}$ will be null independently of the $K_{u}$.
If the \ac{UGV}'s states do not depend entirely on the precedent states and the actual command, the kinetic energy measured may be higher than the kinetic energy commanded.
In that case, the metric also increased because the interaction between the tire and the terrain results in low traction, which makes it harder to accelerate.   

The second most significant advantage is its limited requirements, which are 1) a recorded command in the body frame, 2) a measurement of the \ac{UGV}'s speed in the body frame based on exteroceptive measurements, and 3) geometric measurements of the vehicle ($b$, $c$, $d$).

\section{\MakeUppercase{Experimental setup}}
\label{sec:experimental_setup}

This section presents the platforms used to acquire data to validate the \ac{DRIVE} protocol and assess the behavior of the unpredictability metric $\rho$ on various terrains.
Furthermore, the dataset acquired is presented, detailing the various environments in which the platforms were deployed, as well as the amount of data acquired.

\begin{figure}
    \centering    \includegraphics[width=1\linewidth]{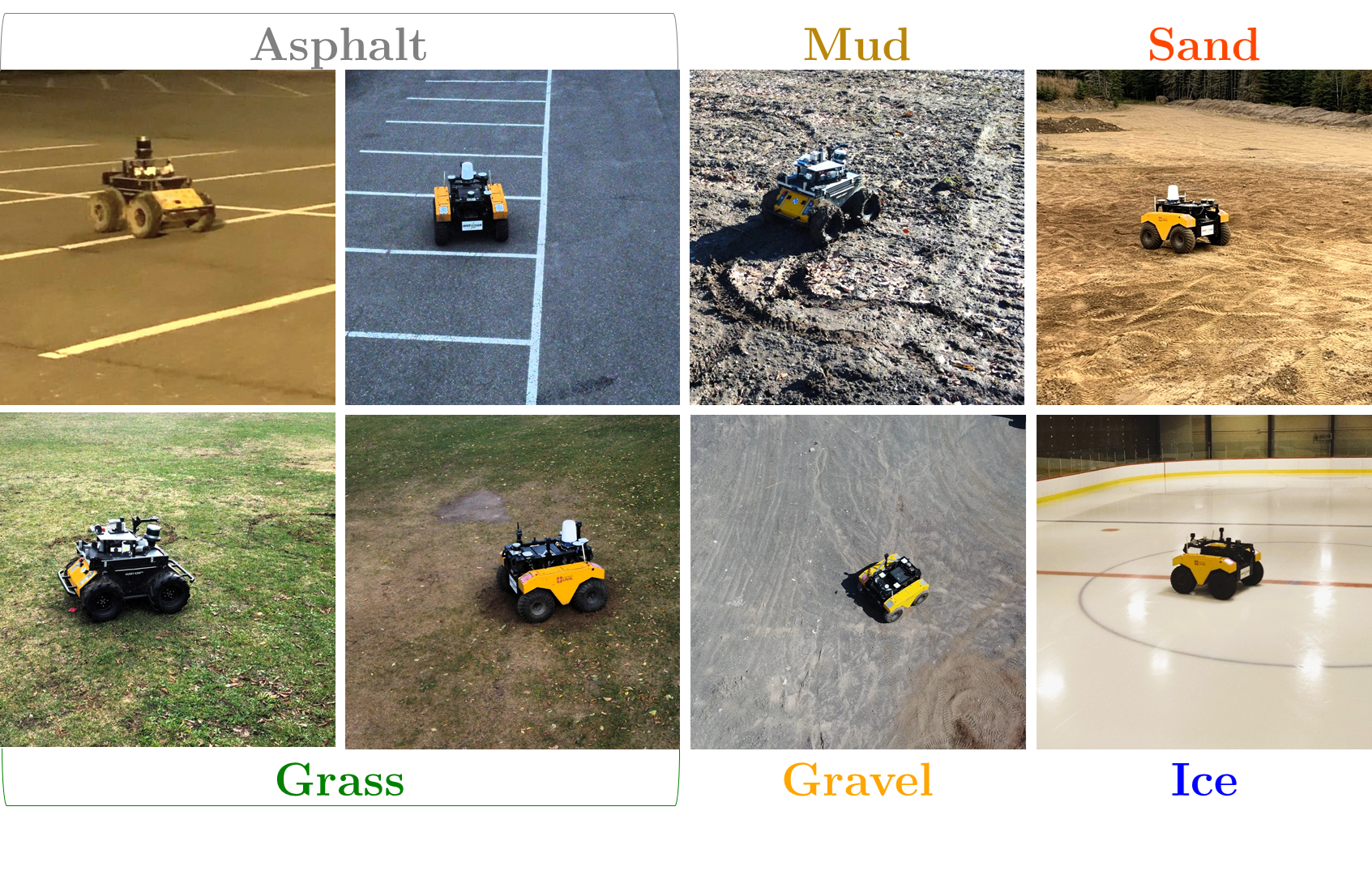}
    \caption{Visual aspect for all terrain and \acp{SSMR} combinations in this experiment. 
    Both asphalt and grass were tested with both the  Clearpath Husky that weighs \SI[detect-weight=true,mode=text]{75}{\kilo \gram} and the Clearpath Warthog that weighs \SI[detect-weight=true,mode=text]{470}{\kilo \gram}.
    }
    \label{fig:allterrain}
\end{figure}

Two distinct \acp{SSMR} were used in the experiments as shown in \autoref{fig:allterrain}.
The first platform is a \textit{Clearpath Robotics} Warthog, weighing $m=\SI{470}{\kg}$ with a top speed advertise as $\dot{t}_{x\text{max}} = \SI{5}{\metre\per\second}$, which was deployed on five different terrains totaling \SI{12.97}{\kilo\metre} and \SI{3.6}{\hour} of continuous driving.
The second platform is a \textit{Clearpath Robotics} Husky, weighing $m=\SI{75}{\kg}$ with a top longitudinal speed of $\dot{t}_{x\text{max}} = \SI{1}{\metre\per\second}$, deployed on three terrains for a total of \SI{1.74}{\kilo\metre} and \SI{1.2}{\hour} of continuous driving.

These two platforms were selected to showcase the difference in vehicle properties for both the analysis of the \ac{DRIVE} protocol and the unpredictability metric.
The experiments conducted and the selected terrains for both platforms are detailed in \autoref{table:dataset} and the visual aspect terrains are shown in \autoref{fig:allterrain}.
The terrain types \emph{Grass} and \emph{Asphalt} were used with both platforms, giving us a common ground to evaluate the impact of similar terrains on different vehicles.

Describing the exact terrain composition is a challenge, but here is an overview of our observations.
The \emph{Mud} terrain was a grass field recovered by \SI{60}{\cm} of dirt. 
The experiments were performed after three days of rain so the dirt became mud. 
The \emph{Sand} terrain was a compacted soil made of sand, forestry byproducts, and dirt. 
The \emph{Gravel} terrain was a hard terrain made of compacted dirt and gravel, similar to the unpaved country roads. 
Finally, the \emph{Ice} terrain was a freshly resurfaced interior ice rink used for skating.

As for the state estimation, the evaluation required to know the wheel velocities $\vecbold{x_{\omega}}$, rotational velocity $\vecbold{x_{\dot{\theta}}}$, and the linear velocities $\vecbold{x_{\dot{t}}}$.
The wheel velocities were estimated by encoder measurements provided by both \acp{SSMR}.
The rotational velocities from an Xsens \ac{IMU} were used as a direct proxy to $\vecbold{x_{\dot{\theta}}}$.
Finally, both \acp{SSMR} are equipped with lidars used in an \ac{ICP}-based localization pipeline \citep{Pomerleau2013} from which we extracted linear velocities.
Our complete experimental dataset totals over \SI{14.71}{\kilo\metre} and \SI{4.9}{\hour} of driving across six terrains and two platforms.

\begin{table}[htbp]
\centering
\caption{Key statistics of the different experiments.} 
\begin{tblr}{Xccccc} 
\cline[1pt, black]{-}
\textbf{Terrain} & \textbf{Platform} &  
 \makecell{\textbf{Number} \\\textbf{of steps}} & \makecell{\textbf{Distance} \\\textbf{travelled (km)}} & \makecell{\textbf{Total} \\\textbf{time (min)}} \\
\cline[1pt, black]{-}
\textcolor{mud}{Mud} & Husky & 162 & 0.47 & 26.0 \\
\textcolor{grass}{Grass} & Husky & 200 & 0.63 & 23.4 \\
\textcolor{grass}{Grass} & Warthog & 200 & 2.14 & 36.2 \\
\textcolor{asphalt}{Asphalt} & Husky & 200 & 0.64 & 23.8 \\
\textcolor{asphalt}{Asphalt} & Warthog & 251 & 2.91 & 39.1 \\
\textcolor{gravel}{Gravel} & Warthog & 220 & 2.50 & 39.6 \\
\textcolor{sand}{Sand} & Warthog & 300 & 3.26 & 44.4 \\
\textcolor{ice}{Ice}  & Warthog & 357 & 2.16 & 59.0 \\
\cline[1pt, black]{-}
& \textbf{Total} & \textbf{1890} & \textbf{14.71} & \textbf{291.5} \\
\cline[1pt, black]{-}
\end{tblr}
\label{table:dataset}
\end{table}

\section{\MakeUppercase{Results}}
\label{sec:results} 

To support our proposed scientific contributions, we evaluated the impact of different terrains on reachable velocities of \acp{SSMR} using the \ac{DRIVE} protocol to validate its utility in terms of system identification.
Moreover, we used the collected data to identify different behaviors of a half-tone robot on complex terrain.
Finally, we validated our unpredictability metric for different terrains and showed its utility in a risk management scenario.

\subsection{\MakeUppercase{Impact of terrain on reachable velocities}}

\begin{figure}[htbp]
    \centering
    \includegraphics[width=\linewidth]{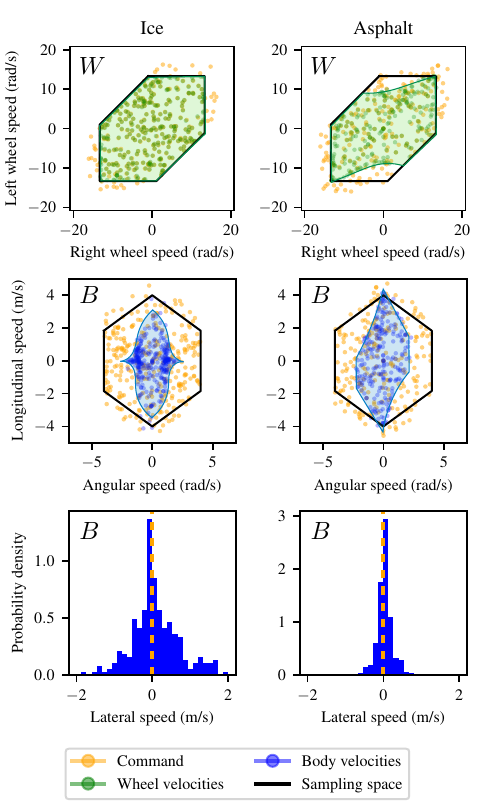}
    \caption{Comparison between the command space $\boldsymbol{\mathcal{U}}$ within black borders, the sampled space $\boldsymbol{\mathcal{S}}$ in orange and the measured steady-state velocities $\boldsymbol{\mathcal{T}}$ over the last two seconds of each command sampled.
    The closed shape in orange in the wheel frame $\boldsymbol{W}$ and in blue for the body frame $\boldsymbol{B}$ are visual approximations of what the continuous space $\boldsymbol{\mathcal{T}}$ is.
    The top row is the identification result in the wheel frame $\boldsymbol{W}$, while the two bottom rows are in the body frame $\boldsymbol{B}$.} 
    \label{fig:command_space_validation}
\end{figure}

This section validates the coverage of the DRIVE protocol over the state space $\vecbold{x}$ of the \ac{UGV}, with a focus on the wheel speed on both sides of the vehicle ($\omega_l$, $\omega_r$) and the velocity of the \ac{UGV} in its body frame ${}^{B}\vecbold{\dot{p}}$.
After sampling commands in $\mathcal{U}$, we analyzed the reachable velocity space $\mathcal{T}$ to assess how much certain types of terrain influence the motion of an \ac{SSMR}.
This validation can be done in the wheel frame $W$ and the body frame $B$. 
As an example, \autoref{fig:command_space_validation} presents the set of sampled commands $\mathcal{S}$ in orange along with the measured speeds in steady state in the wheel frame $W$ (i.e., green points on the top row) and in the body frame $B$ (i.e., blue points in the second row) for the Warthog on ice and asphalt.
Starting with the recorded information in the wheel frame $W$, we observe that the commanded wheel speeds are outside the actual limit of the \ac{SSMR}.
This threshold adjustment is caused by the wheel radius $r$ changing compared to the original parameter used in the \ac{IDD} model.
Indeed, the Warthog has very low tire pressure to allow an all-terrain grip.
This pressure varies through time with extra payloads and operating temperature.
In our case, the wheel radius provided by the manufacturer overestimated the real radius causing saturation of the maximum reachable wheel speed.
Nonetheless, we can observe a major difference between ice and asphalt. The wheels on the ice can cover the full command space such that $\mathcal{U} = \mathcal{T}$.
However, this is not the case for asphalt where we observe an asymmetry in the results.
Due to some wear, the left motor is known to be weaker and is not able to reach its commanded value already at the wheel level.
When focusing on the resulting motion in the body frame $B$ (i.e., the second row of \autoref{fig:command_space_validation}), we observe opposite behaviors.
On ice, although the wheels reach their commanded velocities, the resulting motion in the body frame is heavily reduced.
This reduction is caused by the low friction coefficient between the ice and the tires, which reduces the traction force and, thus, the resistance to the wheel's rotation. 
When comparing the command space $\mathcal{U}$ reduction to the realized velocity space $\mathcal{T}$ in terms of percentage of area, we see that the ice can only reach a small percentage of the total reachable command space.
For terrain with more traction, most of the command space lost is due to the design of the \ac{SSMR} that turns only by slipping, which requires more force than the vehicle's motor has. 
\ac{DRIVE} also identifies well the possible lateral speed of a robot-terrain combination.
One should recall that the lateral command of a \ac{SSMR} is null due to the non-holonomic design of \ac{SSMR}.
However, lateral speed will be produced during high slippage turns of a \ac{SSMR}. 
The last row of \autoref{fig:command_space_validation} shows the normal distribution of lateral speed obtained with the Warthog.  
We can see that the protocol explores well the lateral slip dimension with lateral speed on the ice going as high as \SI{2}{\m\per\second}, which is \SI{50}{\percent} of the maximum forward speed.
On high-friction terrain such as asphalt, this lateral slip is reduced by at least a factor of two.

\begin{figure}[htbp]
    \centering
    \includegraphics[width=\linewidth]{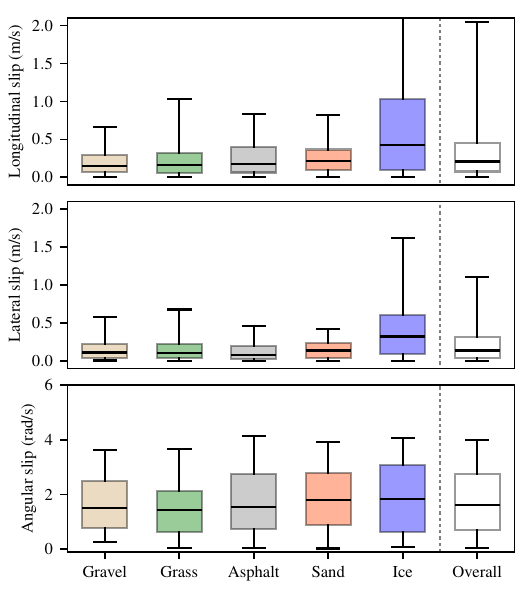}
    \caption{Steady-state slip distributions for the Warthog executing a \ac{DRIVE} protocol on five terrains: asphalt (grey), grass (green), gravel (brown), sand (orange), and ice (blue).
    The right-most box-and-whisker plot in white is the distribution of all terrain combined.
    Each axis represents a dimension of the slip vector in the body frame $\boldsymbol{B}$.}
    \label{fig:slip_boxplot}
\end{figure}

Another use of the \ac{DRIVE} protocol is the investigation of the steady-state slip distribution from a \ac{UGV}-terrain interactions over the command space. 
To be able to fairly compare the terrain, all the distributions used only commands with longitudinal speed between $\pm$\SI{4}{\meter \per \second} and angular speed between $\pm$\SI{4}{\radian \per \second}.  
These slip distributions can then be used to evaluate if multiple terrains similarly impact the \ac{UGV}'s motion. 
Focusing on the Warthog over five terrain types, \autoref{fig:slip_boxplot} shows the distributions for the resulting longitudinal, lateral, and angular slips.
For all of our box-and-whisker representations, we follow the suggestion of \citet{Cumming2009}.
Therefore, the boxes represent the first and the third quartile, while the line inside the box represents the median. 
The whiskers represent \SI{95}{\percent} of the distributions to simplify the interpretation of the box-and-whisker plots. 
This convention is maintained for all box-and-whisker plots throughout the paper and allows us to assess a beginning of significant difference when the median is outside the interquartile range~\citep{Cumming2009}.
Based on this assessment, we can state that all the terrain, except ice, are similar although a small trend can be observed in angular slip.
Similar trends were observed with the Husky having median angular slips of \SI{0.735}{\radian\per\second} on mud, \SI{0.708}{\radian\per\second} on grass, \SI{0.690}{\radian\per\second} on asphalt, all within their respective interquartile distances.
This observation comes as a surprise as many prior works aimed at identifying specific terrain, but we need to recall that we are only observing the Warthog-terrain interaction and not solely the terrain characteristics.
The Warthog being a heavy off-road vehicle, one can assume that an average kinematic motion model, excluding ice, would result in an efficient controller.
As for the iced surface, the measured slip is overshadowed by the dynamics of the vehicle, not its kinematic.
The resulting slip is inflated by large prior commands combined with the high inertia of the Warthog.
The terrain types were hand-picked to be highly different (e.g., soft, hard, rough, different particle sizes), but the small difference between the slip distribution shows well how human-based labeling of the terrain is not adapted to evaluate the difficulty of a terrain for a given \ac{UGV}. 

\subsection{\MakeUppercase{Use case: identification of motion models for the Warthog}}

Having identified the steady-state slips for multiple terrain in the last section, we can compute the transfer function mapping any command in $\mathcal{U}$ to a resulting velocity space $\mathcal{T}$ on a given terrain.
These transfer functions were computed from \ac{DRIVE} protocols deployments and can be used in a motion model to compute the additive slip~\citep{Baril2024}.
Based on our results of the last section, we will focus on two terrain types: ice and asphalt, the latter having similar slips as the other terrain types.
To explicit the deformation of the command space, a Gaussian kernel smoother as described in the section \autoref{sec:theory:slip_analysis} has been convoluted over a mesh grid with a resolution of $0.1$ for the commanded longitudinal speed $u_{\dot{t}x}$ and the commanded rotational speed $u_{\dot{\theta}}$. 
The covariance matrix $\vecbold{\Sigma}$ has been selected to ensure the presence of at least four points within one standard deviation on average.
This constrain resulted in $\vecbold{\Sigma} = 0.64 \vecbold{1}$, where $\vecbold{1}$ is the identity matrix. 

\begin{figure}[htbp]
    \centering
    \includegraphics[width=\linewidth]{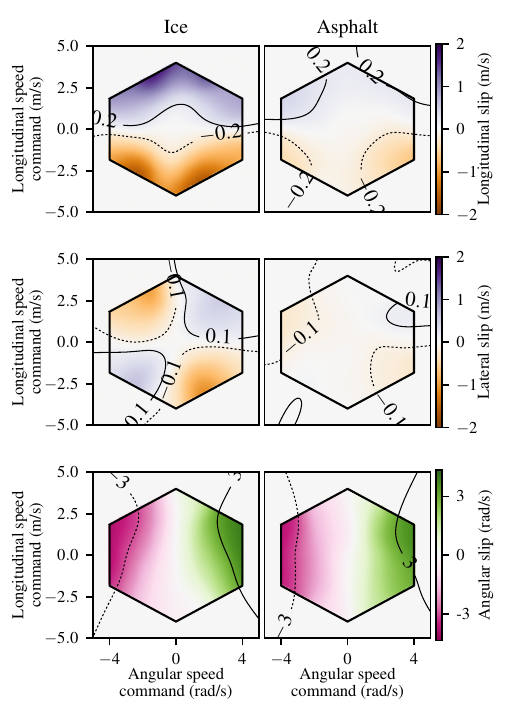}
    \caption{Transfer function between the command space $\boldsymbol{\mathcal{U}}$ in the body frame $\boldsymbol{B}$ of a Warthog robot on ice and asphalt.  
    Contour lines were added to help compare the transfer function for ice (left column) and asphalt (right column).
    } 
    \label{fig:motion_distortion}
\end{figure}

The uniform discrete transfer functions are presented in \autoref{fig:motion_distortion} where axes correspond to commanded speeds (angular and longitudinal), and the color intensity expresses the slip.
On the first row, we see the longitudinal slip $g_{\dot{t}x}$ for the whole command space $\mathcal{U}$.
The transfer function shows that longitudinal slip highly depends on both the angular and longitudinal commands.
The commands that produce the highest longitudinal slip are the commands that maximize both the angular and the longitudinal commands. 
For low-traction combinations such as Warthog-ice, the longitudinal slip mainly depends on the longitudinal command because the vehicle can not reach a longitudinal speed higher than \SI{2.5}{\meter \per second} in the six seconds the sampled command is maintained. 
For a higher-traction terrain-robot combination such as Warthog-asphalt, the longitudinal command does not prevail because the traction between the tires and the terrain is high enough to reach the full longitudinal command space.
To better highlight the difference between low and high traction terrain, the contour lines corresponding to a slip of \SI{0.2}{\meter \per \second} have been added. 

The middle row shows the transfer functions between the command space $\mathcal{U}$ and the lateral slip $g_{\dot{t}y}$.
One should recall that this dimension is not directly controlled.
We can see that lateral slip is practically null when pure angular or linear motion is commanded. 
The lateral slip reaches its peak when both high longitudinal and angular commands are applied. 
These commands result in a vehicle motion commonly called \emph{drifting}, where the rear tires lose more traction than the front tires.
The command to obtain the same lateral slips depends on the \ac{UGV}-terrain traction.
A lower traction \ac{UGV}-terrain combination like Warthog-ice results in larger lateral slips for the same command than the Warthog-asphalt combination. 
To help visualize it, the contour lines associated with a lateral slip of \SI{0.1}{\meter \per \second} are added to the transfer function. 

In the last row, we show the transfer functions between the command space $\mathcal{U}$ and the angular slip $g_{\dot{\theta}}$.
We can observe that the angular slip depends almost uniquely on the angular speed command. 
This is coherent with the design of the \ac{SSMR} that relies on slip to turn. 
The increase of the slip with the angular command shows that the \ac{UGV}-terrain combination established a maximum angular speed, which in our case is above half of the maximum angular speed expected by the \ac{IDD} model.
This behavior is similar for both the Warthog-ice and Warthog-asphalt interaction. 

\subsection{\MakeUppercase{Evaluation of the unpredictability metric}}
In this section, the proposed metric is evaluated on eight datasets collected with the two \acp{SSMR} on six different terrains. 
The metric is evaluated on its consistency within the same \ac{SSMR} on multiple terrains, on its dependency on the command space, and on its ability to quantify the risk associated with \ac{UGV} deployments on a given terrain. 

\subsubsection{Unpredictability metric against terrain types}
\begin{figure}[htbp]
    \centering
    \includegraphics[width=\linewidth]{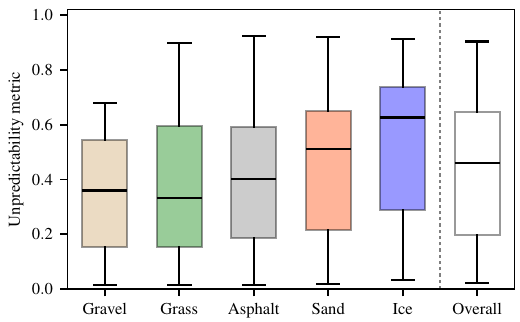}
    \caption{Distributions of the unpredictability metric for the Warthog driving over five terrains. 
    The metric is computed on the steady-state velocities. 
    The colors represent different terrain types and the distribution combining all of them is in white.
        }
    \label{fig:metric_boxplot}
\end{figure}

The first evaluation of the metric aims at verifying the consistency over multiple terrains as well as the relations described in the previous sections. 
Instead of dealing with slip in multiple dimensions, \autoref{fig:metric_boxplot} rates all the terrain tested with the Warthog using a single scalar.
Therefore, the unpredictability metric based on the kinetic energy is coherent with our prior analysis of slip, with ice standing out from grass, gravel, and asphalt for both the unpredictability metric and the slip distributions. 
The median of ice is 1.6 times the median of grass, gravel, and asphalt combined. 
The Warthog-sand combination is the second combination with the highest slip distributions and also the second most unpredictable. 
A continuum seems to appear with the median unpredictability metric of each terrain slightly increased from gravel to ice.
However, similar to the slip distribution, no terrain is significantly different based solely on the unpredictability metric and a \SI{95}{\percent} confidence interval.

\subsubsection{Relation between the unpredictability metric and the command space} 

\begin{figure}[htbp]
    \centering
    \includegraphics[width=\linewidth]{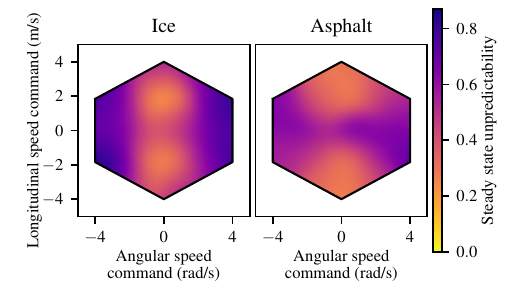}
    \caption{Uniform discrete transfer function between the uncertainty metric and the whole command space $\boldsymbol{\mathcal{U}}$.
    The metric highlights the difference between the steady-state measured velocities and the predicted velocities based on an \ac{IDD} model for the Warthog.
    On the left, are the results for ice.
    On the right, are the ones for asphalt.} 
    \label{fig:metric_command_unpredictability_metric}
\end{figure}

The unpredictability metric can be used to compare the complexity of autonomous navigation in different types of terrain as a whole, but more insights on the behavior of a \ac{UGV} can be understood by analyzing the unpredictability metric over the command space $\mathcal{U}$.
Using the data recorded through \ac{DRIVE}, the unpredictability metric is computed using all the steady-state measured velocity (i.e., on the mean of the last two seconds of each sample) against the predicted velocity from the \ac{IDD} motion model. 
We use the Kernel Gaussian Smoother method described in \autoref{sec:theory:slip_analysis} with the same parameters to interpolate a uniform discrete transfer function over the whole input space $\mathcal{U}$ based on the sample set $\mathcal{S}$. 
The relations between the command space and the unpredictability metric for the deployments with the Warthog on a resurfaced ice rink and asphalt are presented in    \autoref{fig:metric_command_unpredictability_metric}. 
The axes correspond to commanded speeds (angular $u_{\dot{\theta}}$ and longitudinal $u_{\dot{t}}$). 
The color intensity represents the unpredictability metric in the steady state. 
We can observe that the most predictable commands for both terrains are linear motions around $\pm \SI{1.5}{\m \per \second}$. 
The low friction coefficient between the ice and the tire stops the \ac{SSMR} from moving at high speed, which explains that the high-linear speeds are more unpredictable on ice than on asphalt.
As for rotational speed,  more extreme speed increases the unpredictability for both terrains.

Turning on the spot ($u_{\dot{\theta}} \gg u_{\dot{t}}$) is easier on ice because of the low-friction surface. 
Interestingly, the Warthog is harder to predict at low speed.
Indeed, we observe that the suspension and the low-pressure tires act as a torsion spring, which breaks the kinematic assumption of the \ac{IDD} model.
Indeed, for all the terrains tested with the Warthog, turning on the spot produced a minimum unpredictability value of 0.39  and increased with the angular speed command.

This representation shows that the \ac{UGV}-terrain interaction is more complex than only comparing median values for all terrain. 
It is easier to observe differences at the level of the command space $\mathcal{U}$.
This type of mapping has better potential to identify the terrain on which the robot is navigating by identifying the subsamples representing the largest differences among the terrains.  

\subsubsection{Unpredictability metric as a risk assessment tool}

Another advantage of our unpredictability metric is that it can assess the risk level of a given navigation behavior in a specific environment, akin to the risk matrix used in project management~\citep{Cox2008}.
Risk matrices are often a five-by-five grid with the axis representing the \emph{likelihood} and the \emph{severity} of a risk.

On one side, the measured kinetic energy of a vehicle moving relates to how much damage it will cause if it collides with the environment.
This measure can be used as a proxy for the \emph{risk severity}.
The maximal theoretical kinetic energy generated by a vehicle can be computed over the command space using an \ac{IDD} model.
One should note that if external forces are applied to the vehicle (e.g., gravity aligns with the motion of the vehicle, collision with a moving obstacle), this limit can be exceeded.
On the other side, the unpredictability metric $\rho$ relates to the capacity of being able to predict the motion of a robot.
This measure can be used as a proxy for the \emph{risk likelihood}.

\begin{figure}[htbp]
    \centering
    \includegraphics[trim={0 1px 0 0},clip, width=\linewidth]{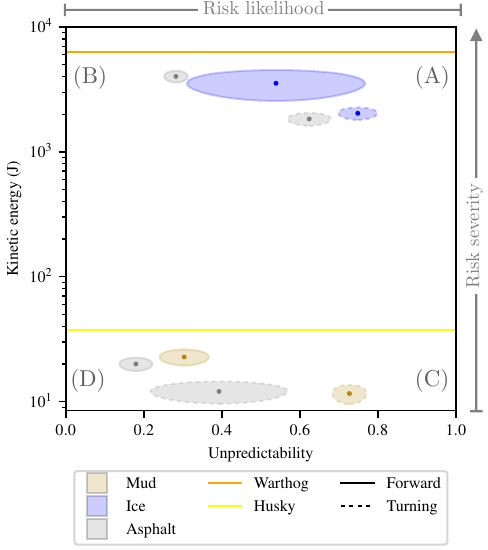}
    \caption{Analysis of \ac{UGV} behaviors using our unpredictability metric and the measured kinetic energy.
    Color dots represent the median value of different navigation scenarios, with their color representing three types of terrain investigated (i.e., asphalt, ice, and mud).
    Their surrounding ellipses encompass \SI[detect-weight=true,mode=text]{95}{\%} of each distribution.
    Ellipses with a solid line represent \emph{motion 1} (i.e., high forward velocities).
    Ellipses with a dashed line represent \emph{motion 2} (i.e., high rotational velocities).
    The same representation can be used to assess risk level (i.e., from low to critical) with extreme examples represented with the letters (A), (B), (C), and (D).
    Horizontal orange and yellow lines represent the maximum kinetic energy for each \ac{UGV} on flat ground (i.e., without external force).
    } 
    \label{fig:metric_deployments}
\end{figure}

The risk of navigation is defined as the combination of these two measures and a continuous version of a risk matrix for autonomous driving is presented in \autoref{fig:metric_deployments}.
In that graph, the behavior of a vehicle with a critical risk level (A) is when it goes at a high speed while being unable to predict where it is going.
A vehicle at high speed, but that can predict its motion offers a moderate level of risk (B) as a collision is improbable.
Similarly, a vehicle barely moving, but unable to predict its motion offers also a moderate risk level (C) as as the severity of a collision will be low.
Finally, a vehicle not moving and not being able to predict so will offer the lowest risk level (D).

Consequently, moving an autonomous vehicle inherently means accepting a certain risk level.
This tolerance to risk is highly dependent on the social context and applications.
For example, researchers investigating controller algorithms will tolerate critical risk given proper safety measures are in place, while a company deploying robot taxis will have little margin from the public perspective.
Nonetheless, our representation allows us to put these different scenarios in perspective.
Moreover, this representation addresses some shortcomings of the risk matrix pointed out by \cite{Cox2008}.
Namely, we are using a continuous axis instead of bins and removing human biases in the qualitative evaluation.

To highlight the advantages of this risk representation, we analyzed further the behavior of two types of motion (forward, turning) on three different terrains (i.e., asphalt, ice, and mud) and for two vehicle sizes (i.e., Warthog and Husky).

The resulting risk assessment is overlaid in \autoref{fig:metric_deployments}.
Starting with the bottom of the graph (i.e., low-risk severity), we see results from the Husky with a combination of low weight (\SI{75}{\kg}) and low maximum velocity.
\emph{Motion 1} (i.e., ellipses with solid lines) explores the resulting risk of driving forward at maximum speed on asphalt and in mud.
Even using a rather crude \ac{IDD} motion model, the \ac{UGV} can well predict its forward motion on asphalt while increasing unpredictability on mud due to slippage.

Lower in the graph, \emph{Motion 2} (i.e., ellipses with dashed lines) explores the resulting risk of turning at maximum speed.
Following the equation for kinetic energy, turning produces less energy than moving forward.
Moreover, turning in mud caused dirt to pile on each side of the wheels, leading to extremely high friction, almost stalling the motors.

At the top of the graph, we see the results for a much larger \ac{SSMR} (i.e., the Warthog weighting \SI{470}{\kg}) at traveling at higher speeds, thus having the potential of causing significantly more damage than the Husky.
Following the same scenarios, the two highest ellipses with solid lines are the results of the Warthog driving at maximum linear speed on asphalt and ice.
The weight of the Warthog, combined with backlash in its suspension resulted in more unpredictable motion when compared to the Husky on equivalent terrains.

Furthermore, driving the Warthog at high speed on the ice is riskier than driving it on asphalt.
The Warthog \emph{motion 1} on ice is mostly governed by its dynamics (i.e., involving acceleration and the high inertia of the \ac{UGV}), breaking many assumptions of the quasi-steady state of a \ac{IDD} model leading to a wide range of unpredictability (i.e., the blue ellipse).
As for \emph{Motion 2}, the Warthog has insufficient tire friction to turn on ice.
Its wheels are spinning fast while generating little motion at the body level, leading to more unpredictability but less variability than for \emph{motion 1}. 

The Husky has the opposite behavior when turning in the mud as its wheels almost stalled instead of spinning.
However, this also results in a significant increase of unpredictability between Motion 1 and 2 as the friction is too high to achieve the command.

On the asphalt, the Warthog has a similar behavior to the Husky in the mud as its tires generate a lot of friction when turning at high speed.
This leads to a significant increase of unpredictability between motions 1 and 2 as Warthog does not have enough power to overcome this friction.
As for the Husky on the asphalt, the increase of the unpredictability between motions 1 and 2 is not as significant as the friction is less of a limiting factor while turning.
Interestingly, this seems to increase the unpredictability variability of \emph{motion 2} as the command's success is more sensitive to tire-friction variations.

\section{\MakeUppercase{Lessons Learned}}
This section regroups a series of advice and lessons learned on the field while deploying the \ac{DRIVE} protocols over \SI{14.7}{\km}. 
The first section contains directions on how to avoid mechanical wear. 
The second section presents technical suggestions on how to deploy \ac{DRIVE} safely and efficiently.
The third subsection presents a further analysis of the \ac{DRIVE} protocol for future potential applications.   

\subsection{\MakeUppercase{Mechanical wear}}
Limiting the maximum acceleration is necessary to prevent mechanical wear on the powertrain and the tires. 
DRIVE currently does a random sampling of the speed set. 
This random sampling can produce drastic acceleration and force on the vehicle powertrain. 
After \SI{13}{\km} of recording with the Warthog with a maximum acceleration of \SI{4}{\meter \per \square \second}, premature mechanical wear was observed and heard. 
By the end of the deployment, skipping gear sounds started to happen with high-torque commands.  
The transmission chain also oscillates and produces noise when the vehicle is commanded to go in the opposite direction of its inertia.

\begin{figure*}[htbp]
    \centering
    \includegraphics[width=0.8\linewidth]{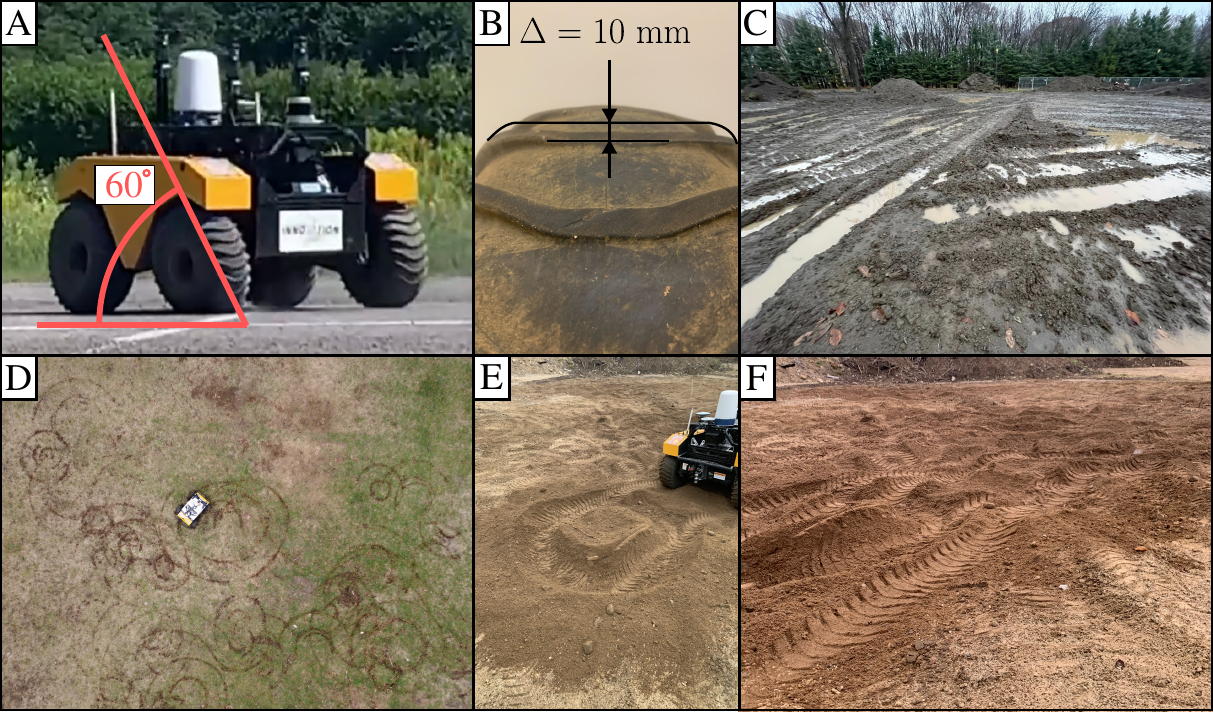}
    \caption{Mosaic of the impact of \ac{DRIVE} on the vehicle and the environment. A) Tire bending due to the aggressive command and the high traction between the tire and the asphalt. B) Thread wear after doing the \SI[detect-weight=true,mode=text]{13}{\kilo \meter} deployment. Most of the wear happened during the 251 sampled commands on asphalt. C) Mud terrain is only homogeneous in small areas, which increases the need for human intervention to keep the \ac{SSMR} on the same type of mud and, therefore, the time to gather data.  D) \ac{DRIVE} protocol tends to remove the grass when turning, even with a small \ac{SSMR} that weighs \SI[detect-weight=true,mode=text]{75}{\kilo \gram}. E) Turning on the spot in the sandy dirt results in the robot digging the ground. F) At the end of \ac{DRIVE}, the relatively uniform terrain became textured due to the vehicle's digging during the protocol.}
    \label{fig:diverse_lesson_learned}
\end{figure*}

The strain on the vehicle highly depends on the traction between the terrain and the vehicle.
Thus, adjusting the acceleration limit by terrain or considering all the terrain you want to deploy when setting these acceleration limits is essential.
In our case, asphalt was the worst offender with its friction causing the wheels to bend inward by \SI{30}{\deg} as shown in \autoref{fig:diverse_lesson_learned}-A.
Another element to consider is tire wear on high-traction terrain. 
The \ac{DRIVE} protocol tends to do many turning maneuvers that induce wheel slipping and skidding for an \ac{SSMR}. 
During the deployment, the Warthog tire treads were reduced from \SI{12}{\milli\metre} to \SI{2}{\milli\metre} in the center of the tire as shown in \autoref{fig:diverse_lesson_learned}-B.
The asphalt was the terrain that produced the highest wear on the tire tread with 251 commands. 
To reduce premature tire wear, it is suggested to either adjust the maximal angular speed of \ac{DRIVE} to prevent the command that results in aggressive turns or to avoid performing \ac{DRIVE} on high traction terrain.
If not, \ac{SSMR} operators should control their tire tread depth and have another set of tires as backup. 
These phenomena seem to be dependent on the SSMR level of kinetic energy as they were not observed on the Husky because of its low weight and speed.

\subsection{\MakeUppercase{Deploying DRIVE efficiently}}
\label{sec:lesson_learned:deploy_drive}

To efficiently deploy the \ac{DRIVE} protocol, it is essential to identify the speed thresholds of the \ac{UGV}, select an appropriate terrain, sample enough points, and be aware of the energy consumed by the protocol. 
It is possible that the \ac{UGV} manufacturer does not provide the speed specifications or provides only an interface to control the \ac{UGV} in the body frame $B$ without explicitly giving the motion model used to send commands to each wheel.
In that case, the missing information can be reverse-engineered.
The speed thresholds can be identified in both the body frame and the wheel frame using an unloaded calibration. 
The first step of the unloaded calibration is to disengage the wheels from the motor or raise the vehicle in the air to remove the contact between the tires and the ground. 
The second step is to send a longitudinal command at maximum speed to measure the maximum wheel speed with the encoders. 
The third step is to measure the base width and the wheel radius and use it with the \ac{IDD} to compute the maximum angular speed from the maximal wheel speed. 
The fourth step is to validate that all the identified command space is accessible and that the commands sent are equivalent to the wheel encoder speed passed through the \ac{IDD}. 
If some speed commands are inaccessible, it is either caused by a software speed limit or a difference in the parameters of the \ac{UGV} manufacturer's motion model and the one used to execute the sampling.
Add the limits to your sampling and repeat the fourth step.
Once the commanded speed is the same as the encoder speed passed through the \ac{IDD} motion model over the whole command space, you are ready to deploy \ac{DRIVE}. 

The minimum terrain space to deploy \ac{DRIVE} is an area where at least one dimension is 1.5 times the distance corresponding to the \ac{UGV} traveling at its maximum linear speed for six seconds. The minimum dimension for deploying \ac{DRIVE} is a rectangle of $20 \times$\SI{45}{\meter} for the Warthog and $9 \times$\SI{6}{\meter} for the Husky.  
Once that criterion is met, the larger the terrain is, the more efficient the protocol realization will be, as an operator needs to interrupt the recording if the \ac{UGV} reaches the boundary of the terrain. 
From \autoref{table:lessons}, this tendency can be seen by comparing the percentage of time used to gather data for the deployment on the grass with the Warthog and the Husky. 
The percentage of time data gathering was computed using the idle time (i.e., the \ac{UGV} on the field but not performing the steps) and the total time of the experiment. 
The idle time is composed of human intervention to stop and reorient the \ac{UGV} when it reaches the border of available space.
This intervention consists of 1) interrupting the six-second commands, 2) reorienting the \ac{UGV} to accomplish the interrupted command towards the center of the available space, and 3) resuming the command.  
The Warthog and Husky were deployed on the same grass location, which was identical in width but not in length.
However, due to the command space of the Warthog being 4 times larger in longitudinal speed and 2.5 times larger on rotational speed than the command space of the Husky, the deployment with the Husky was 1.5 times more efficient than the deployment with the Warthog. 

\begin{table*}[htbp]
\centering
\caption{Statistics for all experiments.}
\begin{tblr}{Xcccccccc} 
\cline[1pt, black]{-}
\textbf{Platform} & \textbf{Terrain} & 
\makecell{\textbf{Max linear} \\\textbf{speed (m/s)}} & \makecell{\textbf{Max angular} \\\textbf{speed (rad/s)}} & \makecell{\textbf{Number} \\\textbf{of sampled command}} & \makecell{\textbf{Distance} \\\textbf{travelled (km)}} & \makecell{\textbf{Total} \\\textbf{time (min)}} & \makecell{\textbf{Time data} \\\textbf{gathering}} &
\makecell{\textbf{Physical space} \\\textbf{(\SI[detect-weight=true,mode=text]{}{\square \meter})}} & \\ 
\cline[1pt, black]{-}
Warthog & \textcolor{sand}{Sand} & $5$ & $5$ & $300$ & $3.26$ & $44.4$ & \SI{72.3}{\percent} & $945$\\
Warthog & \textcolor{asphalt}{Asphalt}  & $5$ & $4$ & $251$ & $2.91$ & $39.1$ & \SI{71.3}{\percent} & $1250$\\
Warthog & \textcolor{grass}{Grass} & $5$ & $5$ & $200$ & $2.14$ & $36.2$ & \SI{60.4}{\percent} & $1100$\\
Warthog & \textcolor{gravel}{Gravel} & $5$ & $4$ & $220$ & $2.50$ & $39.6$ & \SI{60.7}{\percent} & $1900$\\
Warthog & \textcolor{ice}{Ice} & $4$ & $5$ & $197$ & $1.02$ & $34.0$ & \SI{62.8}{\percent} & $940$\\
Warthog & \textcolor{ice}{Ice} & $5$ & $4$ & $160$ & $1.14$ & $25.0$ & \SI{64.0}{\percent} & $740$\\
\cline[1pt, black]{-}
Husky & \textcolor{mud}{Mud} & $1$ & $2$ & $162$ & $0.47$ & $26.0$ & \SI{67.4}{\percent} & $355$\\
Husky & \textcolor{asphalt}{Asphalt}  & $1$ & $2$ & $200$ & $0.64$ & $23.8$ & \SI{85.9}{\percent} & $455$\\
Husky & \textcolor{grass}{Grass} & $1$ & $2$ & $200$ & $0.63$ & $23.4$ & \SI{89.5}{\percent} & $645$\\
\cline[1pt, black]{-}
Total & All & - & - & $1890$ & $14.71$ & $291.5$ & - & -\\
\cline[1pt, black]{-}
\end{tblr}
\label{table:lessons}
\end{table*}

The uniformity assumption of the terrain has the impact of increasing the time to execute the \ac{DRIVE} protocol because it reduces the size of the available homogeneous area. 
A good example of this is the deployment of mud with the Husky. 
The muddy terrain was a dirt area wetted with rain, as shown in \autoref{fig:diverse_lesson_learned}-C.
The water accumulated unevenly in the area causing some parts to have significantly more viscous mud than others.
More interruptions were necessary to reorient the \ac{UGV} in the highly viscous mud to respect the protocol's homogeneous hypothesis.
The increased number of interruptions decreases the \ac{DRIVE} efficiency on Husky by a factor of \SI{20}{\percent} compared to other terrains.  

Another aspect to consider when selecting a terrain to deploy \ac{DRIVE} is the damage it causes to the terrain. 
The high number of turning maneuvers tends to destroy a deformable terrain. 
For instance, the execution of \ac{DRIVE} on grass tends to remove patches of grass, while deployment on the sand and the less cohesive ground tends to dig holes, as presented \autoref{fig:diverse_lesson_learned}-D-E. 
As the experience progresses, these deformations would combine to create a textured ground (\autoref{fig:diverse_lesson_learned}-F), making the Warthog bounce on the ground when passing through a texture at high linear speed.
As an example, one of the terrain used to describe the trajectory is a mix of sand and dirt compacted by the passage of a forest truck. 
During the 300-step recording, it was observed that when the robot turned on itself, it slowly removed the ground under its wheel and sank.
However, this sinking motion only resulted in one-time vehicle immobilization, where the belly of the \ac{SSMR} was transferring part of the vehicle's weight directly to the ground, and the left wheels were turning freely.
The \ac{SSMR} could escape immobilization once a sampled linear-speed command was executed. 
This shows well that the \ac{SSMR} could suffer from fatal immobilization in more dangerous terrain like deep snow, where trying to turn can bury the vehicle in the snow to a point where it can't recover \citep{Baril2022}.

The number of commands to draw when deploying the \ac{DRIVE} protocol highly depends on the application. 
After many tests, 150 commands are the minimal sample size to reasonably estimate the transfer function between the command space and the slip dimensions for the Warthog and the Husky on hard terrains. This number is higher in the mud, being a soft terrain, with the Husky. The covariance matrix used for the Warthog is defined as $\vecbold{\Sigma}= 0.8 \vecbold{1}$ and $\vecbold{\Sigma}= 0.25 \vecbold{1}$ for the Husky.
With this number of commands sampled, there are, on average, four sampled commands in one standard deviation of each command of the mesh grid.
The recommendation is to collect at least 150 sampled commands while keeping in mind that a bigger number of sampled commands increases accuracy. 

The limit to the number of steps becomes a trade-off between the energy consumption and the precision of the slip.  
Depending on the battery, the vehicle, the tested command space, and the terrain, \ac{DRIVE} protocol can consume most of the battery. 
Deploying on deformable terrain increased the energy cost of turning maneuvers. 
For example, the Husky could not complete more than 150 steps of \ac{DRIVE} on muddy terrain but has succeeded in 200 steps on grass with a single battery charge. 
The mud increased the resistance to turning and the energy consumed during these motions. 
A more sophisticated sampling method could consider the energy consumed to reduce the energy consumption of the \ac{DRIVE} protocol while still approximating the relation between the command space and the slip. 

One requirement of the \ac{DRIVE} protocol is to have an accurate localization system, estimating the \ac{UGV} pose and velocity in a global frame.
This requirement can be more problematic than expected because the metric and the \ac{DRIVE} protocol depend on the linear velocities, which can not be directly measured as the angular velocities with an \ac{IMU}.
The computation of linear velocities from the measured linear accelerations is complex due to high centripetal acceleration that often variates during the \ac{DRIVE} random sampling. 
These variations are more important during the first four seconds of the command when the vehicle accelerates to reach a steady state. 
The steady state is also less affected by the accuracy of the linear velocities than the transient state because it is possible to use all the points in the steady state regime to average out the noise. 
Based on this observation, it seems that the \ac{DRIVE} dataset could also be used as a dataset to investigate the robustness of state estimation algorithms against drifting maneuvers that break basic motion model hypothesis \citep{Baril2024}, high variation of the centripetal accelerations that influence the \ac{IMU} in aggressive motions~\citep{Deschenes2024}, and some localization with low-amount of features for lidar \ac{SLAM}.

\subsection{\MakeUppercase{Potential limitations of the protocol}}
This section aims to validate if the current version of \ac{DRIVE} could be used in two cases.
The first case is driving in a highly deformable terrain where the precedent positions of the \ac{UGV} can impact the terrain's mechanical properties. 
A good example is deep snow, where each passage of the \ac{UGV} compacts the snow and changes the traction. 
In that case, the position must be included in the search space because the positions are linked to the mechanical properties of the ground. 
The second case is when the \ac{UGV} needs to execute a work in 
a complex environment where it can not navigate in a steady state. 
In that case, the vehicle state space would include the accelerations.   
Thus, we evaluate the ability of \ac{DRIVE} to explore the positions states space along with the acceleration states.   

\begin{figure}[htbp]
    \centering
    \includegraphics[width=0.5\textwidth]{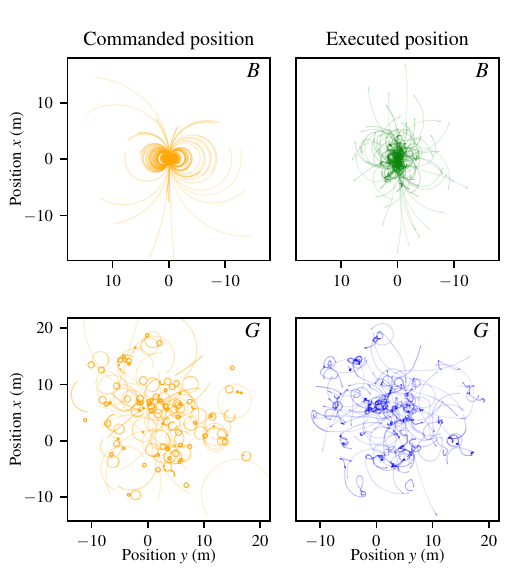}
    \caption{ Analysis of the 177 \ac{DRIVE}-step trajectories with the Warthog on sandy terrain.  
    In the upper row, the commanded trajectories (left) are compared to the measured trajectory (right) both in the body frame $\boldsymbol{B}$. 
    In the lower row, the same comparison, but this time in the global frame $\boldsymbol{G}$.}
    \label{fig:trajectory}
\end{figure}

To evaluate the capacity of \ac{DRIVE} to explore the positions, the commanded trajectories and the measured trajectories of the 177 steps of the Warthog on the sand are presented in the \autoref{fig:trajectory}.
The top row presents the results of the random commands in the body frame $B$.
As we can see from the commanded position, \ac{DRIVE} sampled most of the possible trajectories of an \ac{SSMR} could execute in a six-second window, with an emphasis on rotation over straight lines.
Moreover, the figure on the right presenting the executed trajectories as measured by our localization system shows high discrepancies.
This difference between the commanded and the measured position highlights the unpredictability of the command and the complexity of developing a search algorithm to control the density of the trajectory without a good prior of the motion model.
The bottom row of the same figure shows the same command against the executed trajectories, but this time, it is in the global frame $G$.
We can observe in the global frame measured trajectories that \ac{DRIVE} tends to increase the trajectory density in the center of the map.
This density variation is due to a combination of random sampling and the strategy to resume trajectories when they reach the map borders.
This strategy increases the probability of passing through the center helping to observe the \ac{UGV}-terrain interaction from multiple directions.

\begin{figure}[htbp]
    \centering
    \includegraphics[width=0.5\textwidth]{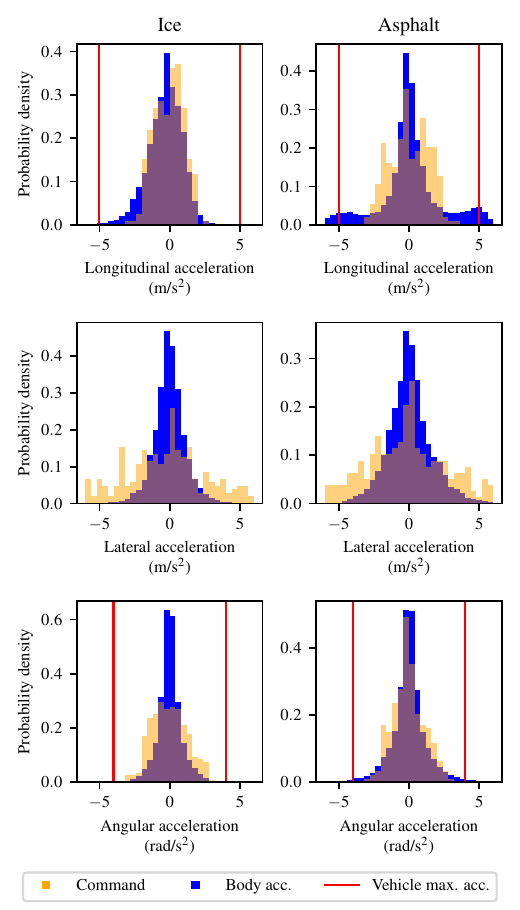}
    \caption{Histogram of the planar acceleration during the first two seconds of the Warthog on ice (left) and on asphalt (right). 
    The manufacturer acceleration limits are presented in red. 
    The theoretical acceleration (orange) is computed from the two seconds of transient state assumption.
    The theoretical lateral acceleration is substituted by the centripetal acceleration caused by the theoretical command.}
    \label{fig:fig:accelerations}
\end{figure}

Although we used the \ac{IDD} motion model in this article, the same acquired dataset could be used to learn the relation between the slip and the command space during the transient state.
The accelerations during the two first seconds of each sampled command cover well transient states. 
As an example, \autoref{fig:fig:accelerations} presents in blue the recorded accelerations of the Warthog on ice and asphalt. 
As we can see, the random-sampling strategy in the velocity space stimulates a large band of accelerations. 
The vehicle's maximal angular and longitudinal accelerations are reached on the asphalt but not the ice.
The lateral acceleration distribution of the asphalt is 1.37 greater than the one on the ice.
\autoref{fig:fig:accelerations} also shows a high difference between the indirectly commanded centripetal acceleration in orange and the measured lateral acceleration in blue.

Random sampling presents two limits concerning the acceleration as it depends on the vehicle speed at the beginning of a sample and the target speed. 
Any constraint that reduces the speed step will bias the acceleration toward smaller values. 
In smaller spaces, the vehicle often needs to be stopped because the vehicle gets frequently too close to available space boundaries. 
Thus, a smaller area tends to reduce the representation of high accelerations as more steps start from zero speed. 
The second limitation is that uniform random sampling on speed results in a triangular probability density function for the speed steps, which presents a positive bias toward low-speed steps and accelerations.

\section{\MakeUppercase{Conclusion}}

To summarize, the \ac{DRIVE} protocol is proposed to the community as a standard procedure to collect data for system identification of \acp{UGV}.
We demonstrated that the protocol succeeds at exploring the velocity states by sampling a subset of commands possible in the wheel frame and the body frame while identifying the longitudinal, lateral, and angular slip distributions of different terrains.
This demonstration was done through \SI{14}{\km} of driving spanning close to five hours of recorded data.
Also, we explored six different types of terrain and observed that there is a discordance between the expectation of a human labeling different terrain and the actual slip measured by a given vehicle.
For example, we showed that human-defined labels such as grass and sand might not be the most relevant to characterize the terrains as they could represent the same unpredictability level for a large \ac{SSMR}.

Replacing the clustering terrain with a characterization metric could ease the comparison of the results of different experiments found in the literature, as we often lack details about the type of ground used in the literature. 
Our unpredictability metric is continuous, and we show that it is sensitive to different terrain compositions.
Moreover, we demonstrated that the collected data enables the estimation of the transfer functions between each dimension of the planar slip and the vehicle command space. 
More specifically, the relationships between the command space and the longitudinal slip, lateral slip, and angular slip have been presented for the Warthog on ice and asphalt.
The results on ice show that the highest longitudinal and lateral slip are measured on commands that maximize both the angular and longitudinal speed, leading to \emph{drifting} maneuvers breaking classical kinematic motion models.
Our proposed unpredictability metric can be used to identify these types of failures.
Furthermore, when combined with the kinetic energy of a \ac{UGV}, it can be used as a risk assessment tool for robotics field deployment.
Therefore, we encourage further publications to report these numbers (i.e., unpredictability metric and kinetic energy) for a given deployment to help compare different experiments.

Some improvements can still be made to the identification protocol.
For example, active path planning could replace manual reorientation towards the center of the map to reduce the trajectory overlap. 
Additionally, active path planning would benefit from online active sampling methods, such as the Gaussian process, to identify better the \ac{UGV}'s limits and sample more uniformly the acceleration to allow the identification of better dynamic modeling.
Finally, we highlighted that it is still challenging to dissociate the \ac{UGV}-terrain interaction into a characteristic that only captures the terrain property.
Such characteristics could be transferred to other \acp{UGV}, thus reducing the need to run the \ac{DRIVE} protocol on multiple systems.

\printbibliography[title={\MakeUppercase{References}}]

\begin{IEEEbiography}
[{\includegraphics[width=1in,height=1.25in,clip,keepaspectratio]{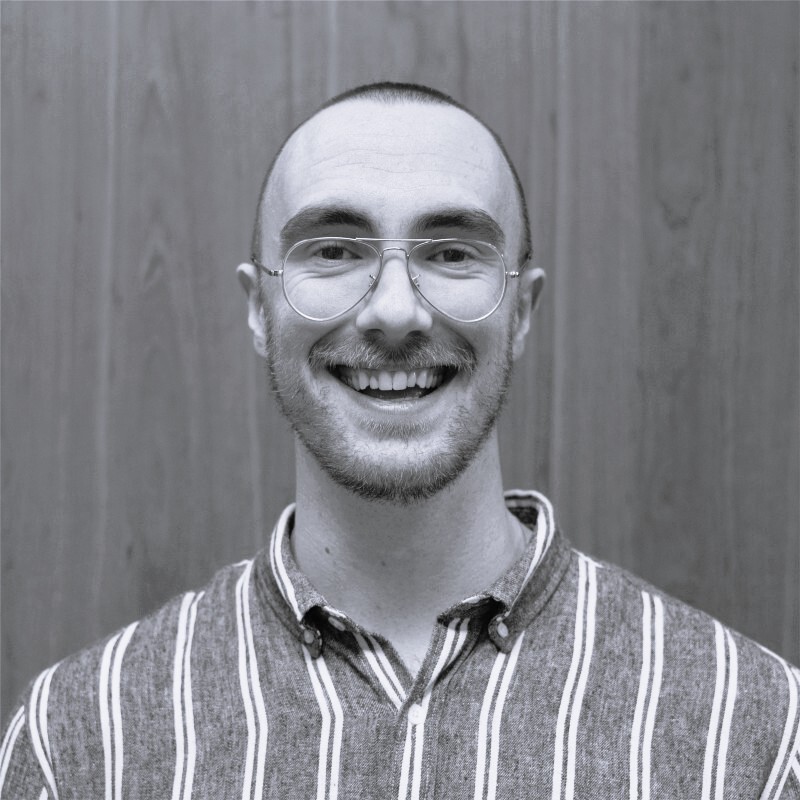}}]
{NICOLAS SAMSON } received the B.Eng. degree in mechatronics, robotics, and automation engineering at Université Laval (Quebec, Canada) in 2023. 

He is currently a Master's student in computer science at Université Laval. His research interests are related to vehicle motion modeling, system identification, and control algorithms.  
\end{IEEEbiography}

\begin{IEEEbiography}
[{\includegraphics[width=1in,height=1.25in,clip,keepaspectratio]{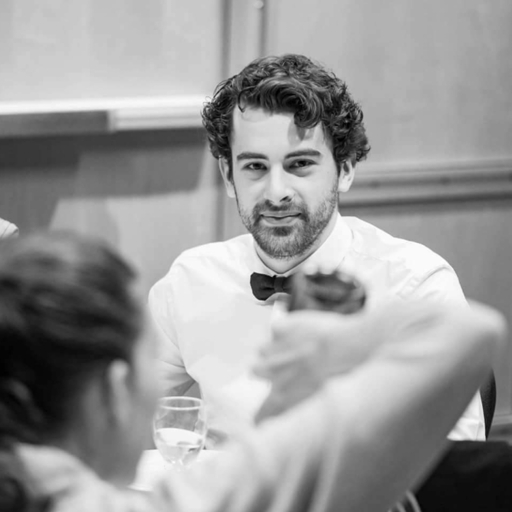}}]
{WILLIAM LARRIVÉE-HARDY } received the B.Eng. degree in computer engineering from Université Laval (Quebec, Canada) in 2023. 

He is currently a Master's student in Computer Science at Université Laval. His research interests are related to deep learning on physical systems to increase autonomy.
\end{IEEEbiography}

\begin{IEEEbiography}
[{\includegraphics[width=1in,height=1.25in,clip,keepaspectratio]{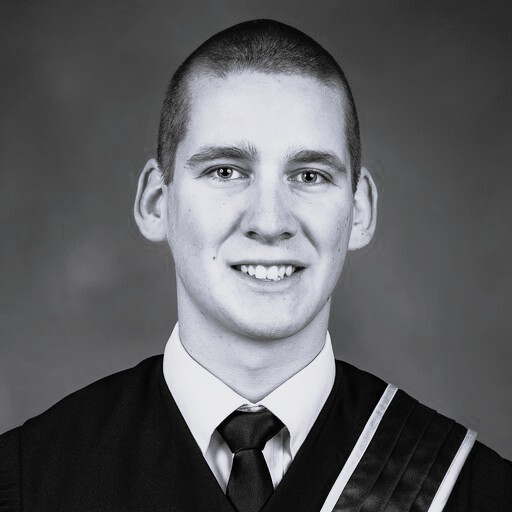}}]
{WILLIAM DUBOIS } received the B. Eng. degree in robotics engineering from Université de Sherbrooke (Quebec, Canada) in 2021. 

He is currently a Ph.D. student in computer science after an accelerated passage from the MS in mobile robotics at Université Laval (Quebec, Canada). His research interests are related to simultaneous localization and mapping, sensor fusion, and system resiliency to harsh weather and localization conditions.    
\end{IEEEbiography}

\begin{IEEEbiography}
[{\includegraphics[width=1in,height=1.25in,clip,keepaspectratio]{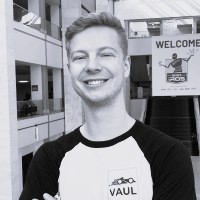}}]
{ÉLIE ROY-BROUARD }  will receive the B.Eng. degree in mechanical engineering from Université Laval (Quebec, Canada) in 2024.

He is currently a B. Eng. student in mechanical engineering at Université Laval. His research interests are related to part design, robotics, and mechatronics.
\end{IEEEbiography}

\begin{IEEEbiography}
[{\includegraphics[width=1in,height=1.25in,clip,keepaspectratio]{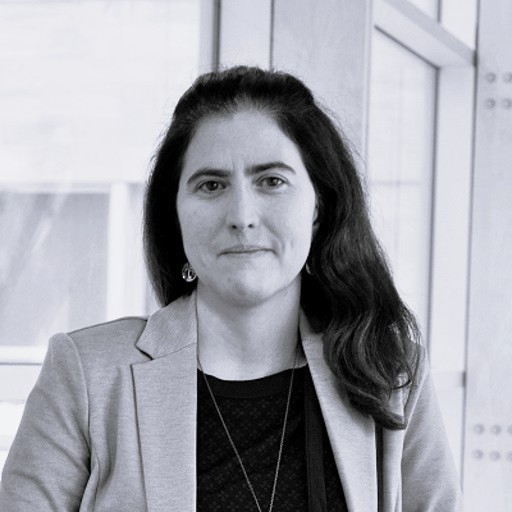}}]
{EDITH BROTHERTON } received an M.B.A. degree in manufacturing and logistics from Université Laval (Quebec, Canada) in 2003 and a Postgraduate diploma in industrial engineering from Université Laval in 2002. 

She has been a scientific research coordinator at the Northern Robotics Laboratory (Quebec, Canada) since 2022. 
Her research interests are related to field deployment and analytics, enabling better mobile robotics systems to be deployed in Northern or difficult conditions. 

\end{IEEEbiography}

\begin{IEEEbiography}
[{\includegraphics[width=1in,height=1.25in,clip,keepaspectratio]{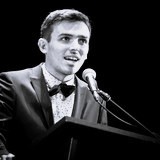}}]
{DOMINIC BARIL } received the Ph.D. degree in computer science from Université Laval (Quebec, Canada) in 2024, and the M.Sc. degree in mechanical engineering from Université de Sherbrooke in 2018.  

He is co-founder and CEO of Tesselate Robotics (Quebec, Canada) with expertise in creating the most robust location and mapping technology available in challenging conditions.  
\end{IEEEbiography}

\begin{IEEEbiography}
[{\includegraphics[width=1in,height=1.25in,clip,keepaspectratio]{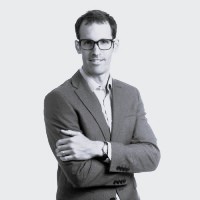}}]
{JULIEN LEPINE } received a Ph.D. degree in mechanical engineering from Victoria University (Melbourne, Australia) in 2016, and a Master's degree in mechanical engineering from Université de Sherbrooke in 2013. 

He is a professor at the Department of Operations and Decision Systems at the Université Laval. His research focus is on sustainable road freight transport including the development of innovative technologies for tires, protective packaging, and freight operation.  
\end{IEEEbiography}

\begin{IEEEbiography}
[{\includegraphics[width=1in,height=1.25in,clip,keepaspectratio]{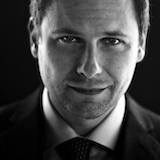}}]
{FRANCOIS POMERLEAU }~(Senior Member, IEEE)~  received a Ph.D. degree in mechanical engineering from the Autonomous Systems Lab at ETH Zurich (Switzerland) in 2013, and the M.Sc. degree in electrical engineering from IntroLab at Université de Sherbrooke in 2009. 

He has been a professor at Université Laval since 2017. His research interests include 3D reconstruction of environments using laser data, autonomous navigation, search and rescue activities, environmental monitoring, trajectory planning, and scientific methodology applied to robotics. 

Prof. Pomerleau is a Professional Member of the Order of Engineering of Quebec (OIQ), a Member of the Research Center of Robotics, Vision and Artificial Intelligence (CeRVIM), and a Member of the Strategic Center of Distributed Intelligent and Shared Environments (REPARTI). 
\end{IEEEbiography}

\vfill\pagebreak

\section{Appendix}

\begin{table}[htbp]
\centering
\caption{Mathematical Notation of key quantities.}
\begin{tblr}{lX} 
\cline[1pt, black]{-}
\textbf{Symbol} & \textbf{Description}
\\
\cline[1pt, black]{-}
$\vecbold{x}, \vecbold{z}, \vecbold{u}$  &
State, observation, and command vectors
\\
$G, B, W$ &
Reference frames for global, vehicle's body, and wheels 
\\
$\mathcal{U}, \mathcal{T}$ &
Command and measured spaces 
\\
$\mathcal{S} = \left\{ \vecbold{u}_1, \vecbold{u}_2, \cdots , \vecbold{u}_s \right\} $ &
Command sampling set with $s$ number of commands
\\
$\vecbold{t} = \begin{bmatrix} t_x & t_y & t_z\end{bmatrix}^T $ &
Translation of the vehicle
\\
$\vecbold{\theta} = \begin{bmatrix} \theta_x & \theta_y & \theta_z\end{bmatrix}^T $ &
Rotation of the vehicle
\\
$\vecbold{p} = \begin{bmatrix} \vecbold{t} & \vecbold{\theta} \end{bmatrix}^T$ &
Pose of the vehicle
\\
$\vecbold{\dot{p}} = \begin{bmatrix} \vecbold{\dot{t}} & \vecbold{\dot{\theta}} \end{bmatrix}^T$ &
Velocities of the vehicle
\\
$\vecbold{\ddot{p}} = \begin{bmatrix} \vecbold{\ddot{t}} & \vecbold{\ddot{\theta}} \end{bmatrix}^T$ &
Accelerations of the vehicle
\\
$\vecbold{\ddot{p}} = \begin{bmatrix} \vecbold{\ddot{t}} & \vecbold{\ddot{\theta}} \end{bmatrix}^T$ &
Accelerations of the vehicle
\\
$\vecbold{\omega} = \begin{bmatrix} \omega_l & \omega_r \end{bmatrix}^T$ &
Wheel velocities on the left and right side
\\
$\omega_\text{max}, \dot{t}_{x\text{max}}, \dot{\theta}_{x\text{max}}$ &
Limits of the command space
\\
$\vecbold{g}$ &
Measured slips expressed as velocities
\\
$\vecbold{\bar{u}}, \vecbold{\bar{g}}$ &
Interpolated commands and slips
\\
$\rho$ &
Unpredictability of a vehicle  
\\
$K, K_{\vecbold{\dot{t}}}, K_{\vecbold{\dot{\theta}}}$ &
Kinetic energy, translational kinetic energy, and rotational kinetic energy  
\\
\cline[1pt, black]{-}
\end{tblr}
\label{table:notation}
\end{table}

\end{document}

%% file: latexGoodPractices/preamble.tex
\usepackage[l2tabu,orthodox]{nag}
\usepackage{amssymb,amsfonts,amsmath,amscd}

\usepackage[
    backend=bibtex8,
    style=ieee,
    sorting=none,
    natbib=true,
    doi=false,
    isbn=false,
    url=false,
    eprint=false,
    maxcitenames=1,
    mincitenames=1
]{biblatex}

\usepackage[pdftex,colorlinks]{hyperref}

\usepackage[french,english,noabbrev,nameinlink]{cleveref}

\usepackage[printonlyused]{acronym}

\usepackage{siunitx}
\usepackage[all]{nowidow}

\usepackage[dvipsnames]{xcolor}

\usepackage{lipsum}

\usepackage{xspace} 

\usepackage[pdftex]{graphicx}

\usepackage{epstopdf}

\usepackage{import}

\graphicspath{{./latexGoodPractices/}}

\usepackage{booktabs}

\usepackage{tabularx}
\usepackage{multirow, multicol}

\usepackage{bm}

\newcommand{\abs}[1]{\left\vert#1\right\vert}

\newcommand{\bbm}{\begin{bmatrix}}
\newcommand{\ebm}{\end{bmatrix}}